\documentclass[sigconf,natbib=true]{acmart}
\AtBeginDocument{%
  }

\usepackage{graphicx}
\usepackage{amsmath}
\usepackage{amsfonts}

\usepackage{booktabs}
\usepackage{subcaption}

\setcopyright{acmlicensed}
\copyrightyear{2025}
\acmYear{2025}
\acmDOI{XXXXXXX.XXXXXXX}
\acmConference[SIGIR '25]{Make sure to enter the correct
  conference title from your rights confirmation email}{July 13-18, 2025}{Padua, Italy}

\begin{document}
\copyrightyear{2025}
\acmYear{2025}
\setcopyright{cc}
\setcctype{by}
\acmConference[SIGIR '25]{Proceedings of the 48th International ACM SIGIR Conference on Research and Development in Information Retrieval}{July 13--18, 2025}{Padua, Italy}
\acmBooktitle{Proceedings of the 48th International ACM SIGIR Conference on Research and Development in Information Retrieval (SIGIR '25), July 13--18, 2025, Padua, Italy}\acmDOI{10.1145/3726302.3730029}
\acmISBN{979-8-4007-1592-1/2025/07}
\title{LUSIFER: Language Universal Space Integration for Enhanced Multilingual Embeddings with Large Language Models}

\author{Hieu Man}
\email{hieum@uoregon.edu}
\affiliation{%
  \institution{University of Oregon}
  \city{Eugene}
  \state{Oregon}
  \country{USA}
}

\author{Nghia Trung Ngo}
\email{nghian@uoregon.edu}
\affiliation{%
  \institution{University of Oregon}
  \city{Eugene}
  \state{Oregon}
  \country{USA}
}

\author{Viet Dac Lai}
\email{viet.lai@adobe.com}
\affiliation{%
  \institution{Adobe Research}
  \city{San Jose}
  \state{California}
  \country{USA}
}

\author{Ryan A. Rossi}
\email{ryrossi@adobe.com}
\affiliation{%
  \institution{Adobe Research}
  \city{San Jose}
  \state{California}
  \country{USA}
}

\author{Franck Dernoncourt}
\email{franck.dernoncourt@adobe.com}
\affiliation{%
  \institution{Adobe Research}
  \city{San Jose}
  \state{California}
  \country{USA}
}

\author{Thien Huu Nguyen}
\email{thienn@uoregon.edu}
\affiliation{%
  \institution{University of Oregon}
  \city{Eugene}
  \state{Oregon}
  \country{USA}
}


\begin{abstract}
Recent advancements in large language models (LLMs) based embedding models have established new state-of-the-art benchmarks for text embedding tasks, particularly in dense vector-based retrieval. However, these models predominantly focus on English, leaving multilingual embedding capabilities largely unexplored. To address this limitation, we present LUSIFER, a novel zero-shot approach that adapts LLM-based embedding models for multilingual tasks without requiring multilingual supervision. LUSIFER's architecture combines a multilingual encoder, serving as a language-universal learner, with an LLM-based embedding model optimized for embedding-specific tasks. These components are seamlessly integrated through a minimal set of trainable parameters that act as a connector, effectively transferring the multilingual encoder's language understanding capabilities to the specialized embedding model. Additionally, to comprehensively evaluate multilingual embedding performance, we introduce a new benchmark encompassing 5 primary embedding tasks, 123 diverse datasets, and coverage across 14 languages. Extensive experimental results demonstrate that LUSIFER significantly enhances the multilingual performance across various embedding tasks, particularly for medium and low-resource languages, without requiring explicit multilingual training data. The code and dataset for training are available at: https://github.com/hieum98/lusifer
\end{abstract}




\maketitle

\section{Introduction}
Text embeddings, which provide dense vector representations of textual content \cite{mikolov2013distributedrepresentationswordsphrases, devlin-etal-2019-bert}, have become fundamental building blocks in modern natural language processing. These embeddings encode semantic information and serve as an important component for numerous downstream applications, ranging from information retrieval and document reranking to classification, clustering, and semantic textual similarity assessment. Recently, the significance of high-quality embeddings has been further amplified by their crucial role in retrieval-augmented generation (RAG) systems \cite{NEURIPS2020_6b493230}. RAG architectures enable large language models (LLMs) to dynamically access and integrate external or proprietary knowledge without the need for model parameter updates, substantially enhancing their adaptability and accuracy \cite{wang2023instructretro, liu2024chatqa, gao2024retrievalaugmentedgenerationlargelanguage}.

The evolution of embedding models has witnessed remarkable advancements, progressing from static word embeddings \cite{robertson2009probabilistic} through contextualized representations \cite{reimers-gurevych-2019-sentence, gao-etal-2021-simcse, ni2021large} to state-of-the-art LLM-based embedding models \cite{wang-etal-2024-improving-text} that harness the sophisticated semantic understanding capabilities of large language models. These developments have substantially enhanced performance across various embedding tasks \cite{luo2024largelanguagemodelsfoundations}, achieving unprecedented accuracy in semantic similarity and retrieval applications. However, a critical limitation remains: the predominant focus on English in LLM-based embedding models has created a significant disparity in multilingual capabilities. This gap is especially pronounced in medium and low-resource languages, where English-centric models exhibit substantial performance degradation due to insufficient language-specific training data \cite{wang-etal-2020-extending, thakur2024leveragingllmssynthesizingtraining}. While recent advances in multilingual embedding models, particularly those leveraging multilingual pre-trained architectures, have demonstrated promising results in multilingual embedding tasks \cite{li2023generaltextembeddingsmultistage, wang2024multilinguale5textembeddings, bge-m3}, their reliance on explicit multilingual supervision for embeddings constrains their applicability primarily to languages with abundant training resources, leaving the challenge of true language-agnostic representation largely unaddressed.

To address this challenge, we present LUSIFER, a novel zero-shot approach that adapts English LLM-based embedding models for multilingual tasks without requiring explicit multilingual supervision. Drawing inspiration from recent advances in multimodal integration \cite{liu2024visual, lu2024ovisstructuralembeddingalignment}, LUSIFER employs a unique architecture that bridges the gap between multilingual understanding and specialized embedding capabilities. At its core, LUSIFER leverages the robust multilingual representations from XLM-R \cite{conneau-etal-2020-unsupervised} and introduces a learnable connector mechanism to interface with English-optimized LLM embedding models. This approach enables LUSIFER to effectively transfer the multilingual understanding of XLM-R to the target LLM while inheriting advanced embedding capabilities of the LLM. In this way, LUSIFER can achieve effective multilingual representation capabilities without requiring explicit multilingual training data. We conduct comprehensive evaluations of LUSIFER through extensive experiments across 123 diverse datasets spanning 14 languages, focusing on five fundamental embedding tasks: Classification, Clustering, Reranking, Retrieval, and Semantic Textual Similarity (STS). Our experimental results demonstrate that LUSIFER substantially enhances the performance of English-centric LLM-based embedding models, achieving average improvements of 3.19 points across all tasks, with particularly significant gains observed for medium and low-resource languages (up to 22.15 improvement). To validate LUSIFER's broader applicability and cross-lingual capabilities, we extend our evaluation to cross-lingual tasks using four comprehensive datasets that encompass over 100 languages, including several critically low-resource languages. LUSIFER significantly outperforms existing English-centric embedding models by 5.75 on average in cross-lingual scenarios. These results demonstrate the effectiveness of our approach in enhancing multilingual representation capabilities without explicit multilingual supervision.

The theoretical foundation for LUSIFER's effectiveness lies in its ability to create a language-agnostic universal space through the integration of a multilingual encoder \cite{pires-etal-2019-multilingual, libovicky-etal-2020-language}. We hypothesize that this universal space serves as a bridge between different languages, enabling the target language model to process semantic information independently of the input language. By mapping these language-neutral representations to the target model's input space, we conjecture that the target LLM can grasp the semantics of these representations, thereby improving the quality of output embeddings across multiple languages. This mechanism allows the model to become less dependent on the specific language of the input, enabling it to better capture semantic information for embedding tasks in languages it rarely encountered during pretraining. Our empirical analysis using t-SNE visualization supports this hypothesis.

\begin{figure*}[t]
  \centering
  \includegraphics[width=\textwidth]{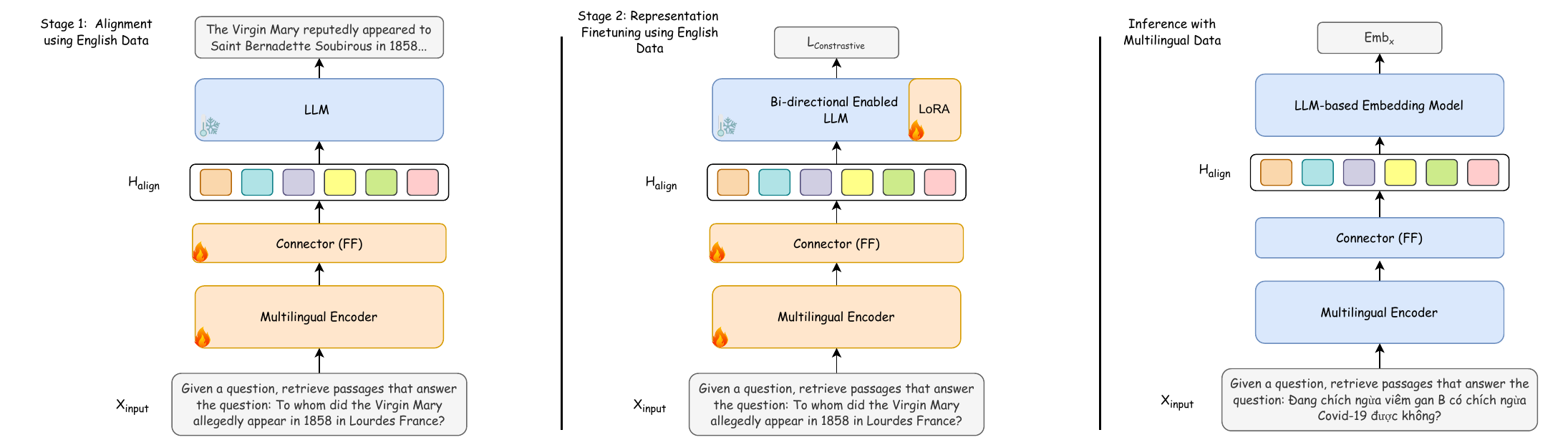}
  \caption{Overview of LUSIFER. \textbf{Left}: Align a multilingual encoder with the target English-centric LLM only using English data and a minimal set of trainable parameter. \textbf{Center}: End-to-end representation finetune through contrastive learning on English text-embedding tasks using LoRA. \textbf{Right}: During inference, LUSIFER successfully processes text-embedding tasks across multiple languages.}
  \label{fig:lusifer-overview}
\end{figure*}

\section{Related Work}
\subsection{English-centric Embedding Models}
Text embedding models have experienced significant advancement in recent years, driven by the evolution of pre-trained language models. Early successes with BERT-based architectures, as demonstrated in Sentence-BERT \cite{reimers-gurevych-2019-sentence}, SimCSE \cite{gao-etal-2021-simcse}, and DPR \cite{karpukhin-etal-2020-dense}, established the foundation for modern embedding approaches. The field has since progressed to leverage LLMs, with recent works \cite{wang-etal-2024-improving-text, muennighoff2024generativerepresentationalinstructiontuning, behnamghader2024llm2veclargelanguagemodels, lee2024nvembedimprovedtechniquestraining, man2024ullmeunifiedframeworklarge} demonstrating substantial improvements in embedding quality and task performance through the enhanced representational capacity of LLMs \cite{luo2024largelanguagemodelsfoundations}.
However, these advances primarily benefit high-resource language applications, as most state-of-the-art LLM-based embedding models are derived from English-centric foundation models \cite{jiang2023mistral7b, touvron2023llamaopenefficientfoundation} and trained predominantly on English or high-resource language datasets \cite{wang-etal-2024-improving-text}. This bias has resulted in a significant performance gap between high-resource and low-resource languages, limiting the global applicability of these models. Our proposed method, LUSIFER, addresses this limitation by enabling effective multilingual representation without multilingual training data.

\subsection{Zero-shot Multilingual Embedding}
Multilingual Embedding has evolved through several distinct methodological approaches, each addressing the fundamental challenge of bridging language gaps in embedding tasks. Early successful approaches relied on translation models to enable multilingual understanding \cite{liu-etal-2020-cross-lingual, shi-etal-2021-cross, zhang-misra-2022-machine}. While effective, these methods introduced operational complexity by requiring external translation systems, limiting their practical deployment and scalability. The emergence of multilingual pre-trained language models, particularly XLM-R \cite{conneau-etal-2020-unsupervised}, opened new possibilities for multilingual transfer. Recent works have demonstrated promising results by fine-tuning such models with contrastive learning objectives on multilingual data \cite{wang2024multilinguale5textembeddings, bge-m3, sturua2024jinaembeddingsv3multilingualembeddingstask}. However, these approaches face two key limitations: they require substantial multilingual training data, and moreover, they do not exploit the sophisticated semantic representations afforded by contemporary English-centric LLM architectures, which have demonstrated superior performance in capturing nuanced semantic relationships.

Recent advances in aligning multilingual and English-centric representations could offer a solution. By combining independently pre-trained representations, a paradigm that has shown remarkable success in multimodal alignment research \cite{NEURIPS2022_960a172b, liu2024visual, lu2024ovisstructuralembeddingalignment}, these works bridge the gap between visual encoders and language models to enhance visual comprehension. As such, similar principles can be applied to align multilingual representations with LLM-based semantic spaces. While related efforts have explored aligning multiple LLMs for improved reasoning capabilities in multilingual settings \cite{bansal2024llm,yoon-etal-2024-langbridge}, these approaches primarily target generation tasks and typically require large-scale alignment data. Our work extends these efforts by focusing on embedding tasks and leveraging a minimal set of parameters to align multilingual and English-centric representations, enabling enhanced multilingual representation capabilities without requirement for large-scale multilingual training data.

\subsection{Multilingual Embedding Benchmarks}
The evaluation landscape for multilingual embedding models has historically been fragmented across various benchmarks, each with significant limitations. While existing benchmarks have made valuable contributions, they often exhibit constrained scope: MINERS \cite{winata2024minersmultilinguallanguagemodels} provides evaluation across multiple languages but is limited to classification and STS tasks with only 11 datasets; XNLI \cite{conneau-etal-2018-xnli}, XQuAD \cite{artetxe-etal-2020-cross}, and SIB-200 \cite{adelani2024sib200simpleinclusivebig} offer broad language coverage but focus exclusively on classification tasks; and MTEB \cite{muennighoff-etal-2023-mteb}, despite its diverse task selection, primarily addresses high-resource languages. To address these limitations, we introduce a comprehensive evaluation framework that encompasses 5 fundamental embedding tasks—Classification, Clustering, Reranking, Retrieval, and STS—across an extensive collection of 123 datasets spanning 14 languages. This holistic approach enables systematic evaluation across both task and language dimensions, providing unprecedented insights into models' multilingual capabilities. Furthermore, our benchmark extends beyond traditional multilingual evaluation by incorporating cross-lingual tasks, featuring coverage of over 100 languages, including critically low-resource languages that have been historically underrepresented in existing benchmarks. This extensive coverage allows for a more nuanced understanding of embedding models' performance across the global linguistic landscape.

\section{Methodology}
Previous works demonstrate that representations of multilingual encoder models exhibit inherent language-agnostic properties, facilitating zero-shot multilingual transfer \cite{pires-etal-2019-multilingual, libovicky-etal-2020-language}. Building upon this foundation, we propose LUSIFER, an embedding framework that aligns a multilingual encoder model with a target English-centric LLM's representational space, enabling the target to encode semantics across multiple languages without extensive multilingual training. This section details our architectural design and two-stage training process for LUSIFER.

\subsection{Model Architecture}
The core development of LUSIFER lies in its novel approach to enabling multilingual encoding of target LLMs through efficient representation mapping. As illustrated in Figure \ref{fig:lusifer-overview}, LUSIFER's architecture consists of three key components: (1) a multilingual encoder that functions as a language-universal learner, capturing semantic information for diverse languages, (2) a language-agnostic connector that serves as a minimal parametric bridge between representations, and (3) a target LLM optimized for embedding-specific tasks. The multilingual encoder processes input from various languages into a shared semantic space, while the connector, designed with minimal trainable parameters, aligns these universal representations with the target LLM's native representational space. This alignment enables the target LLM embedding model to effectively leverage multilingual understanding without requiring extensive multilingual training data or architectural modifications.

Following successful approaches in multimodal alignment \cite{NEURIPS2022_960a172b, liu2024visual, lu2024ovisstructuralembeddingalignment}, we implement the connector as a 2-layers feed-forward network, $\mathbf{FF}$, augmented with a single trainable token appended to the multilingual encoder's hidden states. Formally, given input tokens $\mathbf{X}_{input}$ (with necessary padding), the multilingual encoder's hidden states $\mathbf{H}_{enc}$ are transformed to align with the target LLM's representational space. The resulting aligned hidden states $\mathbf{H}_{align}$ maintain dimensionality compatibility with the target LLM's hidden states while extending the sequence length by one ($|\mathbf{X}_{input}| + 1$):
$\mathbf{H}_{align} = [\mathbf{FF}(\mathbf{H}_{enc}); \mathbf{t}]$, where $\mathbf{FF}$ is the feed-forward network to align the multilingual encoder's hidden states with dimension $\mathbf{d}_{e}$ to the target LLM's hidden states with dimension $\mathbf{d}_{t}$, and $\mathbf{t} \in \mathbb{R}^{d_t}$ is the trainable token. Moreover, we employ a masking mechanism to mask any original padding tokens in $\mathbf{H}_{enc}$ to prevent their influence on the target LLM's processing, ensuring the model focuses on meaningful tokens. 

\subsection{Training Pipeline}
LUSIFER employs a two-stage training process to achieve optimal multilingual representation capabilities. Both stages only require training on English data, leveraging the multilingual encoder's inherent language-agnostic properties and embedding advantages of LLMs to facilitate zero-shot multilingual transfer.

\begin{figure*}[t]
  \centering
  \includegraphics[width=\textwidth]{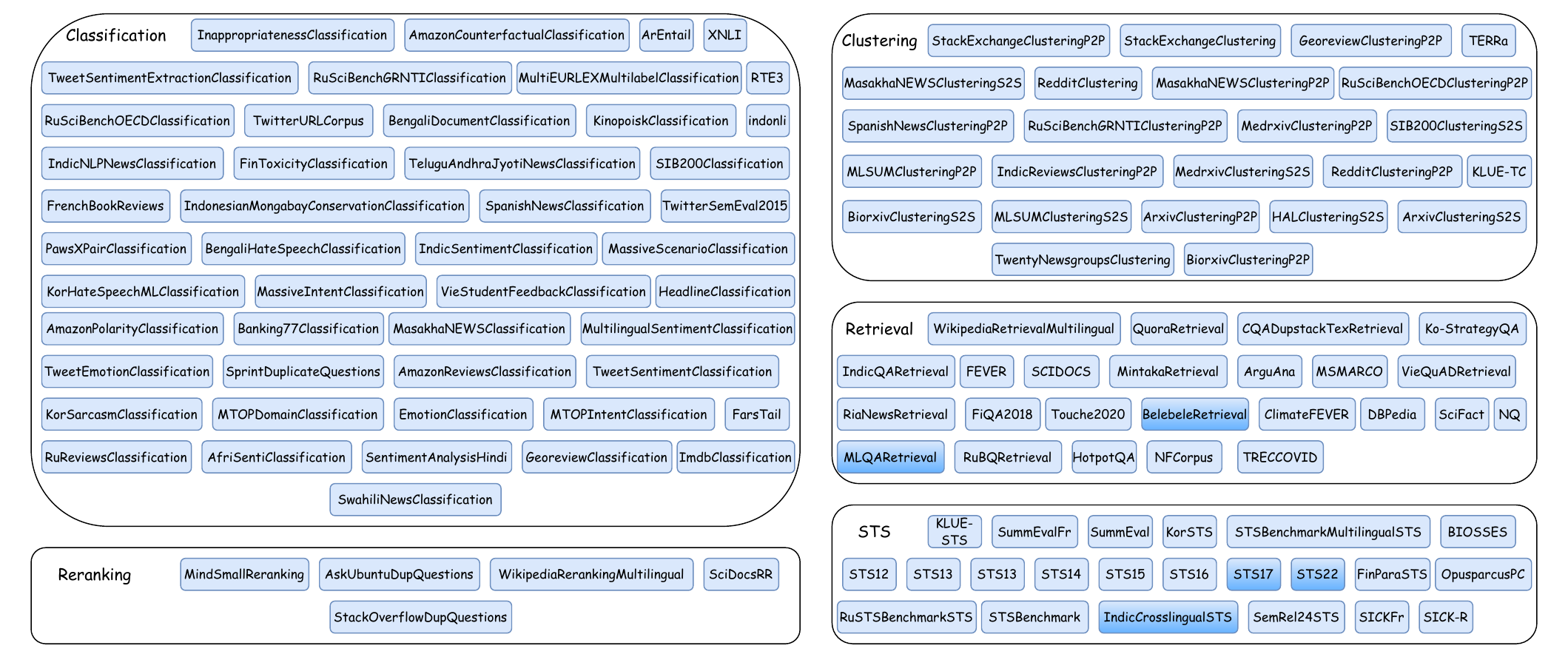}
  \caption{Overview of tasks and datasets in our benchmark. Crosslingual datasets are marked with a blue shade.}
  \label{fig:lusifer-benchmark}
\end{figure*}

\textbf{Stage 1: Alignment Training.}
The initial training stage aligns the multilingual encoder's representations with the target LLM's embedding space. Specifically, we optimize the connector parameters $\theta_c$ and the multilingual encoder parameters $\theta_e$ while keeping the target LLM's parameters fixed, ensuring stable convergence. The training employs two complementary objectives: (1) A masked reconstruction task where we randomly mask $k\%$ of input tokens such that $\mathbf{X}_{input} = \texttt{mask}(\mathbf{X}, k)$, training the model to recover the original sequence $\mathbf{X}_{lm} = \mathbf{X}$. (2) An autoregressive completion task that focuses on next-token prediction, where the model learns to generate the target sequence $\mathbf{X}_{lm}$ conditioned on the input context $\mathbf{X}_{input}$. The training objective for both tasks is formulated as language modeling objective to generate the target sequence $\mathbf{X}_{lm}$ given the input sequence $\mathbf{X}_{input}$. This objective enables local token-level alignment through masked reconstruction task where the model learns to predict the masked tokens by leveraging the context. In addition, it exploits global semantic alignment through autoregressive completion task that encourages the model to capture semantic information of the input sequences to generate the target sequence. As such, our training strategy learns to align the multilingual encoder's representations with the target LLM's embedding space while preserving important semantic information of multilingual input sequences. Our training process is conducted using the standard cross-entropy loss function. This stage aims to establish a strong alignment between the multilingual encoder and the target LLM, enabling the target LLM to effectively process multilingual representations produced by the multilingual encoder.

\textbf{Stage 2: Representation Finetuning.} 
The second stage improves text representations through a contrastive learning process, effectively teaching the model to distinguish between positive and negative examples. Our approach leverages both in-batch negatives sampled from the current training batch and hard-negative examples specifically curated to enhance model training. Additionally, we incorporate bidirectional attention mechanisms within the target LLM, following recent advances in LLM's representation learning \cite{muennighoff2024generativerepresentationalinstructiontuning, behnamghader2024llm2veclargelanguagemodels, lee2024nvembedimprovedtechniquestraining,man2024ullmeunifiedframeworklarge}. This bidirectional context modeling significantly enhances the quality of learned representations by enabling the model to capture both forward and backward dependencies in the input sequence. During this stage, we finetune all components of LUSIFER, including the target LLM, the multilingual encoder, and the connector parameters, to optimize the model's representation quality for embedding-specific tasks. The goal of this stage is to improve the quality of text representations by leveraging the advanced embedding capabilities of the target LLM while maintaining the multilingual understanding provided by the multilingual encoder.

The two-stage training process enables LUSIFER to effectively align multilingual representations with the target LLM's embedding space, enhancing the target LLM's multilingual representation capabilities without requiring explicit multilingual supervision. 

\section{Experiment}
This section presents our experimental methodology and results. We first introduce benchmark datasets and evaluation metrics in Section \ref{sec:benchmark}, followed by our experimental setup including model implementation, training data, and procedures in Section \ref{sec:setup}. We then present our main findings in Section \ref{sec:results} and analyze LUSIFER's cross-lingual capabilities in Section \ref{sec:crosslingual}. We examine detail performance of LUSIFER across five embedding tasks in Section \ref{sec:task_analysis} and evaluate LUSIFER's component effectiveness in Section \ref{sec:analysis}. Finally, Section \ref{sec:visualization} visualizes LUSIFER's representations in multilingual space to demonstrate its language-agnostic capabilities.

\begin{table*}[!ht]
  \centering
  \small
  \resizebox{\textwidth}{!}{
  \begin{tabular}{l|ccccccccccccccc}
      Baselines & En & Es & Ru & Fr & Vi & Fa & Id & Ar & Fi & Ko & Hi & Bn & Te & Sw & Avg. \\ 
      \midrule
      Jina-embeddings-v3* \cite{sturua2024jinaembeddingsv3multilingualembeddingstask} & 59.84 & 61.23 & 62.88 & 58.94 & 66.74 & 78.35 & 58.51 & 64.71 & 73.57 & 64.96 & 64.19 & 61.54 & 68.96 & 49.20 & 63.83 \\ 
      
      mGTE-base* \cite{zhang2024mgte} & 60.40 & 59.65 & 61.02 & 56.20 & 65.81 & 73.46 & 56.55 & 61.97 & 68.96 & 61.22 & 60.81 & 58.24 & 63.58 & 52.57 & 61.46 \\ 

      BGE-M3* \cite{bge-m3} & 60.09 & 60.60 & 62.37 & 57.34 & 70.69 & 78.97 & 58.78 & 64.12 & 75.60 & 64.72 & 64.61 & 65.31 & 69.85 & 54.20 & 64.80 \\

      Multilingual-E5-large* \cite{wang2024multilingual} & 61.91 & 61.97 & 62.91 & 59.40 & 71.30 & 78.08 & 55.21 & 63.41 & 76.53 & 66.55 & 63.75 & 63.67 & 67.32 & 51.55 & 64.54 \\ 
      
      UDEVER-Bloom-7B* \cite{zhang2023language} & 55.83 & 56.39 & 59.73 & 54.38 & 64.32 & 68.70 & 48.97 & 55.02 & 67.60 & 58.54 & 55.96 & 55.13 & 61.00 & 47.41 & 57.78 \\ 
      \midrule

      SimCSE \cite{gao-etal-2021-simcse} & 51.92 & 51.81 & 24.90 & 46.95 & 31.18 & 37.12 & 39.27 & 29.46 & 41.64 & 26.23 & 25.17 & 21.54 & 26.71 & 38.36 & 35.16 \\ 

      Contriever \cite{izacard2022unsuperviseddenseinformationretrieval} & 49.29 & 44.26 & 26.55 & 44.05 & 33.03 & 39.66 & 38.33 & 32.36 & 45.76 & 26.47 & 23.27 & 22.61 & 22.64 & 39.26 & 34.82 \\ 

      GTE-large \cite{li2023generaltextembeddingsmultistage} & 62.29 & 51.66 & 33.49 & 50.13 & 38.88 & 44.67 & 43.07 & 30.27 & 51.98 & 27.02 & 20.38 & 22.97 & 22.75 & 41.40 & 38.64 \\ 

      BGE-en-1.5 \cite{bge_embedding} & 63.27 & 51.65 & 32.79 & 50.84 & 38.50 & 49.73 & 43.28 & 30.81 & 51.16 & 31.11 & 25.28 & 26.34 & 23.02 & 41.96 & 39.98 \\ 

      E5-large \cite{wang2024textembeddingsweaklysupervisedcontrastive} & 60.12 & 52.41 & 26.81 & 51.00 & 37.99 & 39.47 & 43.86 & 31.32 & 53.59 & 28.84 & 24.57 & 23.48 & 22.03 & 43.25 & 38.48 \\ 

      ST5-XXL \cite{ni2021sentencet5scalablesentenceencoders} & 58.81 & 60.35 & 44.42 & 58.50 & 41.81 & 24.66 & 53.43 & 25.30 & 52.46 & 15.43 & 18.07 & 17.10 & 21.63 & 38.81 & 37.91 \\ 

      GTR-XXL \cite{ni2021largedualencodersgeneralizable} & 58.12 & 54.39 & 41.94 & 53.21 & 37.96 & 24.67 & 50.08 & 25.14 & 53.88 & 15.23 & 17.35 & 15.92 & 22.12 & 40.57 & 36.47 \\ 

      E5-Mistral \cite{wang-etal-2024-improving-text} & \bf{66.64} & \bf{61.84} & \bf{61.30} & \bf{59.65} & 58.58 & 72.55 & 58.25 & 54.43 & 66.97 & 62.82 & 56.23 & 55.10 & 47.15 & 50.61 & 59.44 \\ 
      \midrule

      LUSIFER \textbf{(Ours)} & 57.20 & 60.14 & 59.82 & 59.24 & \bf{67.69} & \bf{76.17} & \bf{59.70} & \bf{55.60} & \bf{72.83} & \bf{65.23} & \bf{62.37} & \bf{58.43} & \bf{69.30} & \bf{53.12} & \bf{62.63} \\
      \bottomrule
  \end{tabular}
  }
  \caption{Comparative analysis of model performance across multiple languages and tasks. The table presents average metrics for each model, with the highest score for each language emphasized in bold. * denotes the models trained on extensive multilingual data.}
  \label{tab:results}
\end{table*}

\begin{table*}[!ht]
  \centering
  \small
  \resizebox{\textwidth}{!}{
  \begin{tabular}{l|cccccc}
      Baselines & MLQARetrieval & BelebeleRetrieval & STS17 & STS22 &IndicCrosslingual  & Avg. \\ 
      \midrule

      SimCSE \cite{gao-etal-2021-simcse} & 7.41	& 18.35 & 39.71 &	37.95 &	0.18 & 20.72\\ 

      Contriever \cite{izacard2022unsuperviseddenseinformationretrieval} & 9.75 &	22.94 &	34.55	& 41.72	& 0.03 & 21.80\\ 

      GTE-large \cite{li2023generaltextembeddingsmultistage} & 16.99 & 31.82 &	37.57 &	53.79 & 1.59 &	28.35\\ 

      BGE-en-1.5 \cite{bge_embedding} & 16.64 & 31.19 &	40.40 &	50.77 &	1.11 & 28.02\\ 

      E5-large \cite{wang2024textembeddingsweaklysupervisedcontrastive} & 17.04 &	31.12 &	37.90 &	54.31 &	1.83 & 28.44\\ 

      ST5-XXL \cite{ni2021sentencet5scalablesentenceencoders} & 20.82 &	41.68 &	56.19 &	59.02 &	1.76 & 35.89\\ 

      GTR-XXL \cite{ni2021largedualencodersgeneralizable} & 20.19 & 38.02 &	50.83 &	60.11 &	2.74 & 34.38\\ 

      E5-Mistral \cite{wang-etal-2024-improving-text} & 31.54 &	54.75 &	\bf{81.12}	& \bf{71.37} &	21.92 &	52.14\\ 
      \midrule

      LUSIFER \textbf{(Ours)} & \bf{36.68} & \bf{57.81} &	81.09 &	70.49 &	\bf{43.40} &	\bf{57.89}\\
      \bottomrule
  \end{tabular}
  }
  \caption{Cross-lingual evaluation results. The table presents average metrics for each model over all languages of the datasets, with the highest score for each language emphasized in bold.}
  \label{tab:crosslingual}
\end{table*}

\subsection{Benchmark}
\label{sec:benchmark}
Figure \ref{fig:lusifer-benchmark} illustrates the tasks and datasets in our benchmark. Following \cite{muennighoff-etal-2023-mteb}, our benchmark includes five fundamental embedding tasks, with the evaluation protocol for each task adapted from the respective original papers. The benchmark involves 123 diverse datasets, including 48 Classification datasets, 24 Clustering datasets, 24 Retrieval datasets, 22 Semantic Textual Similarity STS datasets, and 5 Reranking datasets. The main metrics for each task are as follows: Classification: Accuracy, Clustering: V-measure \cite{rosenberg-hirschberg-2007-v}, Retrieval: nDCG@10, STS: Pearson correlation based on cosine similarity \cite{reimers-etal-2016-task}, and Reranking: MAP. Following \cite{lai-etal-2023-chatgpt}, our benchmark covers 14 languages including 5 high-resource languages: English (en), Spanish (es), Russian (ru), French (fr), Vietnamese (vi); 6 medium-resource languages: Persian (fa), Indonesian (id), Arabic (ar), Finnish (fi), Korean (ko), Hindi (hi); 3 low-resource languages: Bengali (bn), Telugu (te), Swahili (sw). 

Additionally, we evaluate models on cross-lingual retrieval tasks where the models need to perform text embedding tasks with queries and documents in different languages. These tasks feature 5 datasets, including Belebele \cite{Bandarkar2024}, MLQA \cite{lewis-etal-2020-mlqa}, STS17, STS22 \cite{agirre-etal-2016-semeval}, and IndicCrosslingualSTS \cite{DBLP:journals/tacl/RameshDBJASSDJK22}, covering over 100 languages, including critically low-resource languages.

\begin{figure*}[t]
  \centering
  \subcaptionbox{Classification tasks}{\includegraphics[width=0.50\textwidth]{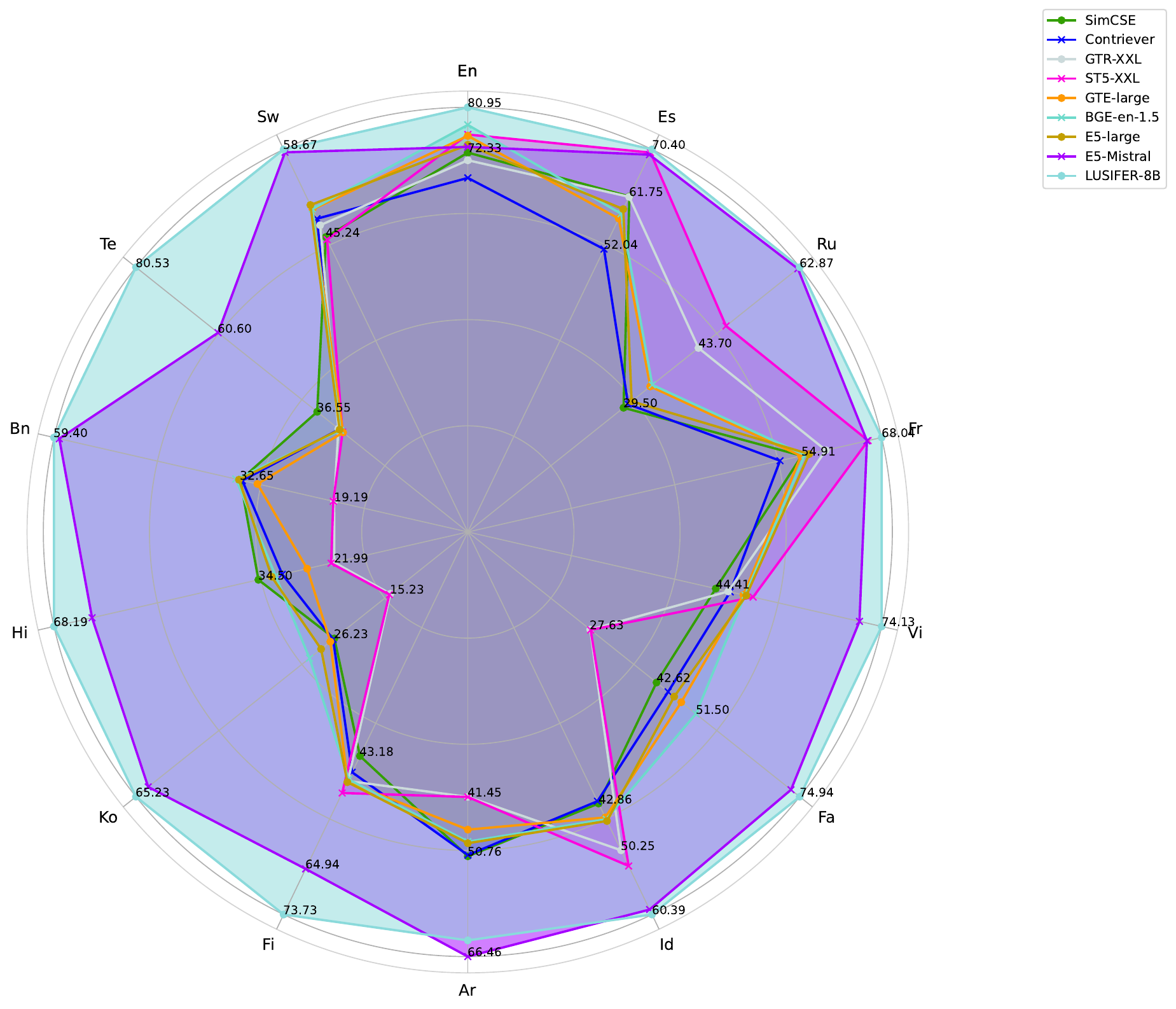}}%
  \hfill % <-- Seperation
  \subcaptionbox{Clustering tasks}{\includegraphics[width=0.50\textwidth]{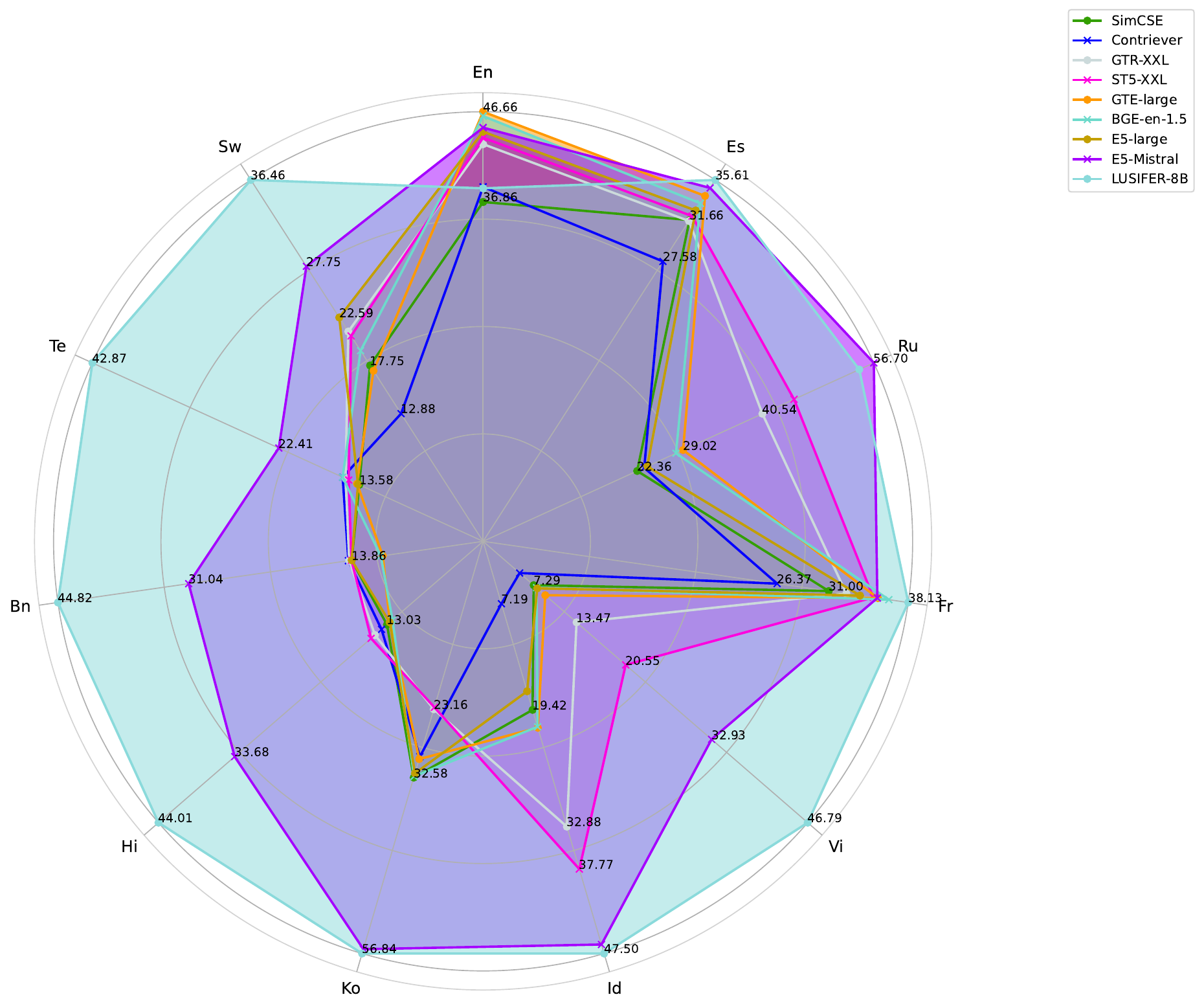}}
  \caption{Performance comparison of LUSIFER and baseline models on Classification and Clustering tasks.}
  \label{fig:classification_clustering}
\end{figure*}

\subsection{Experimental Setup}
\label{sec:setup}
\textbf{Implementation Details.} LUSIFER encompasses three key components: a multilingual encoder, a connector, and a target LLM. We employ XLM-R-large \cite{conneau-etal-2020-unsupervised} as the multilingual encoder, Mistral-7B \cite{jiang2023mistral7b} as the English-centric target LLM, and a 2-layer feed-forward network with one trainable token as the connector. To facilitate efficient training, we leverage the LoRA framework \cite{hu2022lora} for training of LUSIFER's components. Furthermore, we employ GradCache \cite{gao-etal-2021-scaling}, gradient checkpointing, mixed precision training, and FSDP \cite{zhao2023pytorchfsdpexperiencesscaling} to minimize GPU memory requirements. The LUSIFER architecture and its training code are built on top of the Hugging Face Transformers \cite{wolf-etal-2020-transformers} and Pytorch Lightning libraries \cite{falcon_2024_13254264}. We detail the training hyper-parameters for each stage in Table \ref{tab:hprams} 

\begin{table}[t]
  \centering
  \small
  \resizebox{\columnwidth}{!}{
  \begin{tabular}{l|c|c}
    \toprule
    \textbf{Hyperparameter} & \textbf{Alignment Training} & \textbf{Representation Finetuning} \\
    \midrule
    Batch size & 256 & 256 \\
    Learning rate & 1.5e-4 & 5e-5 \\
    Learning rate scheduler & cosine & cosine \\
    Learning rate warm-up ratio & 0.1 & 0.1 \\
    Weight decay & 0.01 & 0.01 \\
    Grad norm clipping & 1.0 & 1.0 \\
    Epochs & 2 & 1 \\
    Optimizer & AdamW & AdamW \\
    Float precision & bf16-mixed & bf16-mixed \\
    LoRA rank & 16 & 16 \\
    LoRA alpha & 32 & 32 \\
    Random mask ratio & 0.5 & - \\
    Number of hardnegatives & - & 7 \\
    \bottomrule
  \end{tabular}
  }
  \caption{Training hyperparameters for each stage.}
  \label{tab:hprams}
\end{table}

\textbf{Training Data.} We only train LUSIFER on a diverse public English datasets. For alignment training, we use the combination of the English Wikipedia and questions-answering datasets. Specifically, we use subset of Wikitext-103 \cite{merity2017pointer} and MSMARCO \cite{bajaj2018msmarcohumangenerated} for the masked reconstruction and autoregressive completion tasks, respectively. For representation finetuning, we adopt the retrieval datasets as follows: MS MARCO \cite{bajaj2018msmarcohumangenerated}, NQ \cite{kwiatkowski-etal-2019-natural}, PAQ \cite{lewis-etal-2021-paq}, HotpotQA \cite{yang-etal-2018-hotpotqa}, SNLI \cite{bowman-etal-2015-large}, SQuAD \cite{rajpurkar-etal-2016-squad}, ArguAna \cite{wachsmuth-etal-2018-retrieval}, FiQA \cite{10.1145/3184558.3192301} and FEVER \cite{thorne-etal-2018-fever}. To address the lack of hard negatives in these datasets, we leverage an encoder-based model \cite{wang2024textembeddingsweaklysupervisedcontrastive} to select the hard negatives on those datasets. Refer to Table \ref{tab:training_data} for the number of samples used in each dataset.

\begin{table}[t]
  \centering
  \small
  \resizebox{\columnwidth}{!}{
  \begin{tabular}{l|l|c}
    \toprule
    \textbf{Stage} & \textbf{Dataset} & \textbf{Number of Samples} \\
    \midrule
    Alignment Training & Wikitext-103 \cite{merity2017pointer} & 100,000 \\
    \text{ } & MSMARCO \cite{bajaj2018msmarcohumangenerated} & 100,000 \\
    \midrule
    Representation Finetuning & MS MARCO \cite{bajaj2018msmarcohumangenerated} & 100,000 \\
    \text{ } & FEVER \cite{thorne-etal-2018-fever} & 100,000 \\
    \text{ } & PAQ \cite{lewis-etal-2021-paq} & 100,000 \\
    \text{ } & SNLI \cite{bowman-etal-2015-large} & 100,000 \\
    \text{ } & HotpotQA \cite{yang-etal-2018-hotpotqa} & 97,800 \\
    \text{ } & SQuAD \cite{rajpurkar-etal-2016-squad} & 97,400 \\
    \text{ } & FiQA \cite{10.1145/3184558.3192301} & 6,420 \\
    \text{ } & NQ \cite{kwiatkowski-etal-2019-natural} & 3,420 \\
    \text{ } & ArguAna \cite{wachsmuth-etal-2018-retrieval} & 1,280 \\
    \bottomrule
  \end{tabular}
  }
  \caption{Number of samples used in each dataset for training. The number of negative samples is included in the total number of samples.}
  \label{tab:training_data}
\end{table}

\textbf{Baselines.} We evaluate LUSIFER's performance across the five fundamental embedding tasks on the benchmark datasets. We make comparisons with a variety of baseline models for embedding tasks which only trained/finetuned on mainly English data. Baselines include the following categories: dense retrieval models with Small Language Model (SLM) backbone: SimCSE \cite{gao-etal-2021-simcse}, Contriever \cite{izacard2022unsuperviseddenseinformationretrieval}, GTE-large \cite{li2023generaltextembeddingsmultistage}, BGE-en-1.5 \cite{bge_embedding}, E5-large \cite{wang2024textembeddingsweaklysupervisedcontrastive}; and dense retrieval models with Large Language Model (LLM) backbone: GTR-XXL \cite{ni2021largedualencodersgeneralizable}, ST5-XXL \cite{ni2021sentencet5scalablesentenceencoders}, E5-Mistral \cite{wang-etal-2024-improving-text}. Moreover, we include the following state-of-the-art multilingual embedding models which are trained on extensive multilingual data for reference: Jina-embeddings-v3 \cite{sturua2024jinaembeddingsv3multilingualembeddingstask}, mGTE-base \cite{zhang2024mgte}, BGE-M3 \cite{bge-m3}, Multilingual-E5-large \cite{wang2024multilingual}, and UDEVER-Bloom-7B \cite{zhang2023language}. 

\subsection{Main Results}
\label{sec:results}

Table \ref{tab:results} presents a comprehensive evaluation of LUSIFER's performance across 14 diverse languages, demonstrating its capabilities in multilingual representation learning. Our model achieves state-of-the-art performance in 10 out of 14 languages, with an average score of 62.63, surpassing the previous benchmark set by E5-Mistral (59.44) by 3.19 points. This improvement is particularly noteworthy given that E5-Mistral utilizes extensive proprietary synthetic data and multilingual training resources. The performance distribution across different language categories reveals LUSIFER's robust multilingual capabilities. In high-resource languages, while maintaining competitive performance with established benchmarks, our model shows particular strength in medium and low-resource languages. The most striking improvement is observed in Telugu (te), where LUSIFER achieves a remarkable 22.15 points gain over E5-Mistral, underscoring its effectiveness in enhancing representation capabilities for traditionally underrepresented languages. When compared to existing embedding models with SLM backbones, LUSIFER demonstrates substantial improvements over models like E5-large (38.48) and BGE-en-1.5 (39.98) which are trained on English data only, thus further demonstrating the benefits of combining multilingual encoder and LLM's English-centric for text-embedding tasks in multilingual settings.

Furthermore, LUSIFER achieves competitive performance (62.63) against state-of-the-art multilingual models such as BGE-M3 (64.80) and Multilingual-E5-large (64.54), despite these models requiring extensive multilingual training data. A key advantage of LUSIFER lies in its resource efficiency. While conventional approaches often rely heavily on extensive multilingual training data, our model achieves comparable or superior performance through its innovative alignment mechanism. This efficiency is particularly valuable in scenarios where multilingual training resources are limited or costly to obtain. The model's ability to generalize effectively across languages while maintaining high performance demonstrates the robustness of our approach.

\subsection{Cross-Lingual Evaluation}
\label{sec:crosslingual}
\begin{figure}[t]
  \centering
  \includegraphics[width=0.50\textwidth]{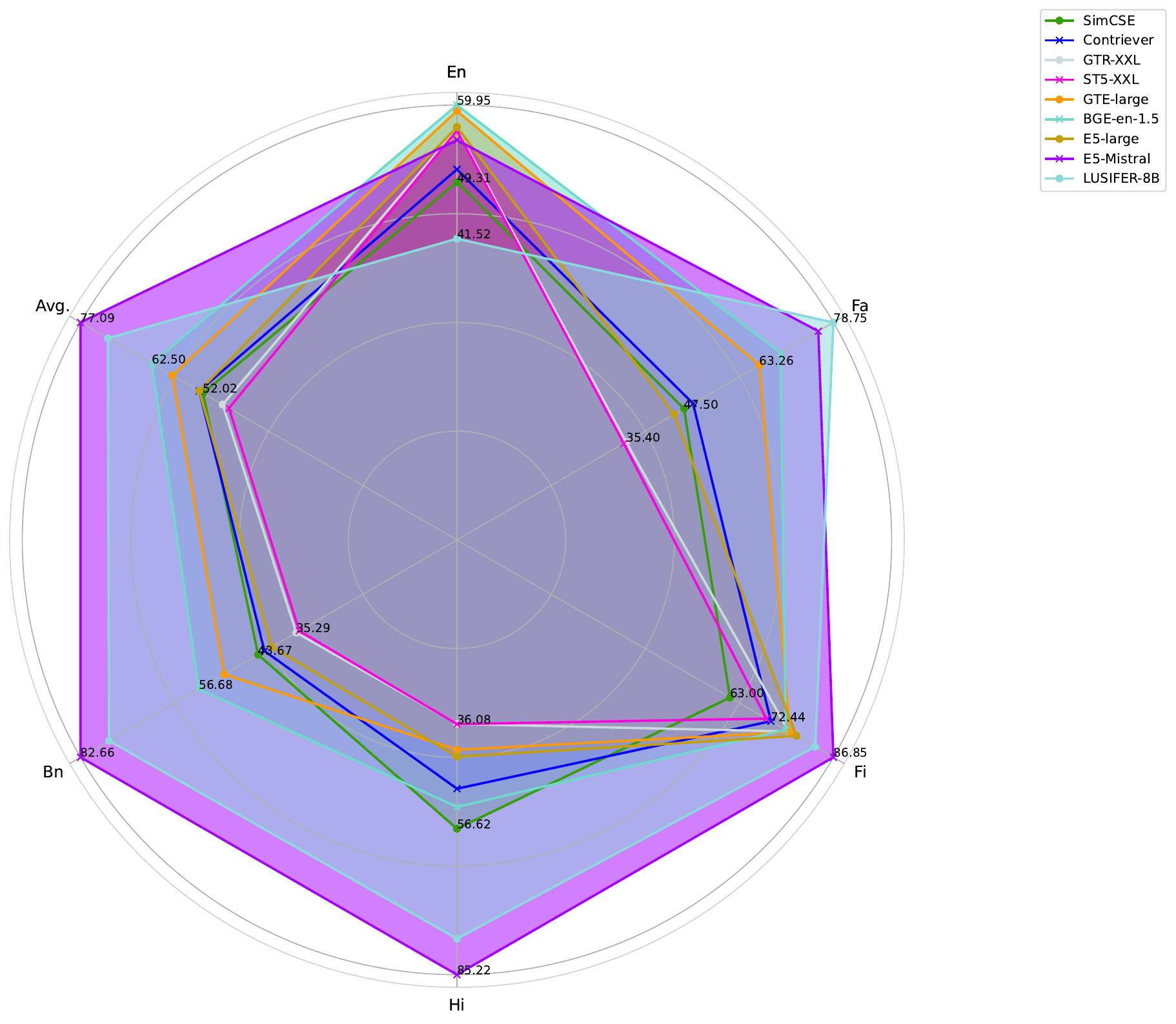}
  \caption{Performance comparison of LUSIFER and baseline models on Reranking tasks.}
  \label{fig:reranking}
\end{figure}

\begin{figure*}[t]
  \centering
  \subcaptionbox{Retrieval tasks}{\includegraphics[width=0.50\textwidth]{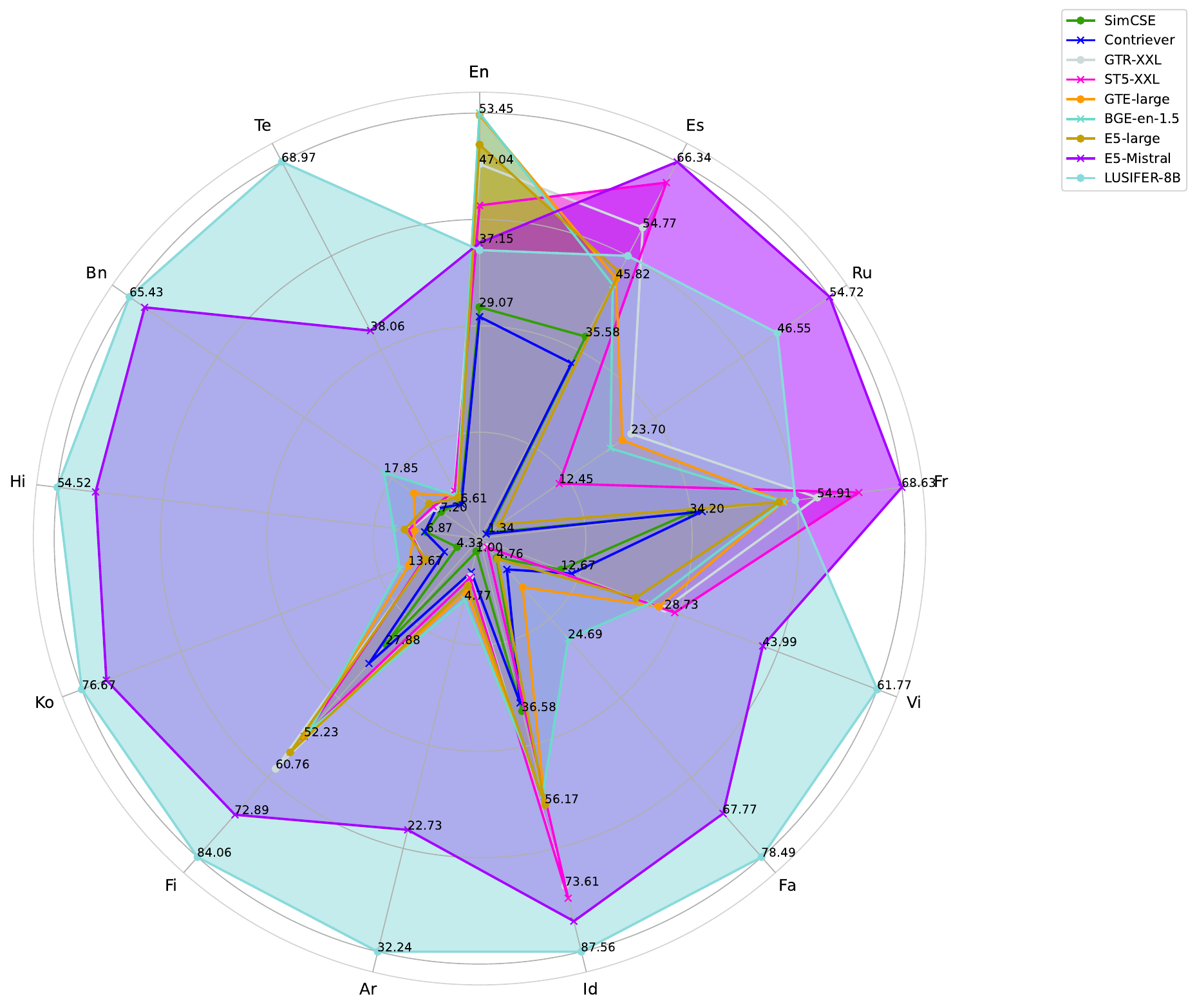}}%
  \hfill % <-- Seperation
  \subcaptionbox{STS tasks}{\includegraphics[width=0.50\textwidth]{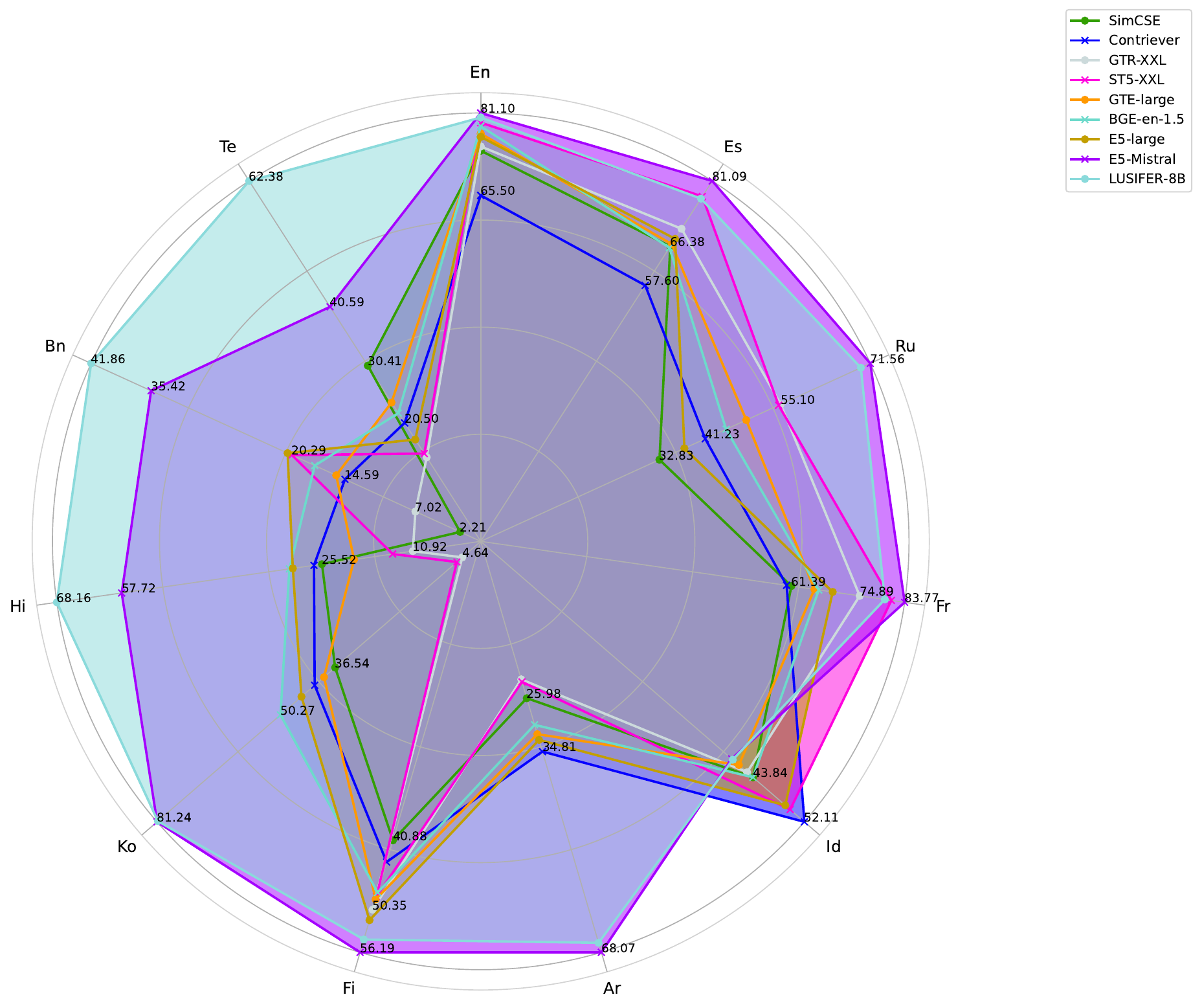}}%
  \caption{Performance comparison of LUSIFER and baseline models on Retrieval and STS tasks.}
  \label{fig:retrieval_sts}
\end{figure*}

LUSIFER demonstrates cross-lingual capabilities across multiple languages and tasks, as evidenced by the evaluation results presented in Table \ref{tab:crosslingual}. Our model achieves a state-of-the-art average score of 57.89, surpassing the previous benchmark set by E5-Mistral (52.14) by a significant margin of 5.75 points. The performance breakdown across different datasets reveals LUSIFER's consistent superiority. In the MLQARetrieval, LUSIFER achieves a score of 36.68, demonstrating a 5.14 improvement over E5-Mistral's 31.54. Similarly, on the BelebeleRetrieval dataset, LUSIFER's score of 57.81 outperforms E5-Mistral's 54.75, showcasing its robust cross-lingual retrieval capabilities. For semantic similarity tasks, LUSIFER maintains competitive performance, achieving scores of 81.09 and 70.49 on STS17 and STS22 respectively, closely matching or slightly trailing E5-Mistral's performance. Perhaps the most striking achievement of LUSIFER is its performance on low-resource languages, particularly evident in the IndicCrosslingual dataset. Here, LUSIFER achieves an unprecedented score of 43.40, nearly doubling E5-Mistral's performance of 21.92. This remarkable improvement demonstrates LUSIFER's ability to effectively transfer semantic knowledge across language boundaries, particularly benefiting languages with limited resources. These results underscore LUSIFER's effectiveness in enhancing cross-lingual capabilities through efficient multilingual representation alignment, enabling the model to process text-embedding tasks across multiple languages effectively. 

\subsection{Task-Specific Performance}
\label{sec:task_analysis}

Figure \ref{fig:classification_clustering}, \ref{fig:reranking}, \ref{fig:retrieval_sts} present the performance comparison of LUSIFER and baseline models on Classification, Clustering, Reranking, Retrieval, and STS tasks. LUSIFER consistently outperforms the baseline models across 4 out of 5 tasks, with the largest improvements observed in Clustering and Retrieval tasks, especially in the medium and low-resource languages. However, the performance of LUSIFER in the Reranking tasks is slightly worse than the strongest baseline, E5-Mistral model. This discrepancy may be attributed to the task's complexity and the information loss in the alignment process between the multilingual encoder and the target LLM. Nevertheless, LUSIFER's strong performance across a variety of tasks and languages highlights its ability to enhance multilingual representations without relying on explicit multilingual training data.

\begin{table*}[!ht]
  \centering
  \small
  \resizebox{\textwidth}{!}{
  \begin{tabular}{l|ccccccccccccccc}
      Baselines & En & Es & Ru & Fr & Vi & Fa & Id & Ar & Fi & Ko & Hi & Bn & Te & Sw & Avg. \\ 
      \midrule
      LUSIFER \textbf{(Full)} & \bf{57.20} & \bf{60.14} & \bf{59.82} & \bf{59.24} & \bf{67.69} & \bf{76.17} & \bf{59.70} & \bf{55.60} & \bf{72.83} & \bf{65.23} & \bf{62.37} & \bf{58.43} & \bf{69.30} & \bf{53.12} & \bf{62.63} \\

      \midrule

      LUSIFER \textbf{(Connector Only)} & 35.53 &	33.98 &	42.95 &	33.54 &	35.68	& 57.86 &	35.55 &	27.60 &	48.72 &	34.45 &	47.57 &	41.85 &	46.50 &	34.66 &	44.18 \\

      LUSIFER \textbf{(Frozen Multilingual Encoder)} & 50.99 & 58.77 & 58.30 & 52.73 & 62.24 & 75.88 & 58.11 & 41.66 & 70.75 & 59.53 & 62.48 & 55.53 & 66.24 & 49.12 & 58.74 \\

      LUSIFER \textbf{(Alignment Only)} & 43.32	& 38.94	& 45.12 &	36.75 &	41.96 &	64.60 &	38.38 &	33.07 &	52.78 &	38.08 &	53.06 &	47.84 &	48.34 &	40.03 &	44.45 \\
      
      LUSIFER \textbf{(Representation Finetuning Only)} & 49.71 & 58.76 & 58.08 &	51.01 &	62.11 &	74.01 &	57.32 &	40.95 &	68.47 &	57.81 &	59.74 &	53.53	& 63.39 &	47.03 &	57.28 \\
      \bottomrule
  \end{tabular}
  }
  \caption{Ablation study results of LUSIFER's components. The table presents average metrics for each model, with the highest score for each language emphasized in bold.}
  \label{tab:ablation}
\end{table*}

\subsection{Ablation Study}
\label{sec:analysis}

To comprehensively evaluate LUSIFER's architectural design and training methodology, we conducted an extensive ablation study examining the contribution of each major component. We compared LUSIFER against several ablated variants:

\begin{enumerate}
    \item \textbf{Connector-Only}: A simplified version using only the finetuning connector during both alignment training and representation finetuning stages, removing the complex interaction between components.
    
    \item \textbf{Frozen Multilingual Encoder}: A variant where the multilingual encoder is freeze throughout training, with only the connector trained during alignment and both connector and target LLM trained during representation finetuning.
    
    \item \textbf{Alignment Only}: The model trained solely with alignment training, omitting the representation finetuning stage to assess the importance of the two-stage training process.
    
    \item \textbf{Representation Finetuning Only}: A version trained using only representation finetuning without initial alignment training.
\end{enumerate}

\noindent We evaluated these variants across our standard benchmark suite, measuring performance on cross-lingual transfer, semantic similarity, and downstream task effectiveness. Table \ref{tab:ablation} presents the comprehensive results of this analysis. The full LUSIFER model achieved the highest average score of 62.63, significantly outperforming all ablated versions. Breaking down the performance impacts: (i) The Connector-Only variant (44.18) showed a 18.45 point, highlighting the importance of allowing the multilingual encoder and the target LLM to be finetuned during training. (ii) The Frozen Multilingual Encoder variant (56.74) performed better than Connector-Only but still fell short by 3.89, demonstrating the value of end-to-end training. (iii) Alignment-Only (44.45) and Representation Finetuning Only (57.28) variants both showed substantial degradation, with drops of 18.18 and 5.35 respectively, indicating the complementary nature of our two-stage training approach, especially the importance of the representation finetuning stage. Through this ablation study, we can attributed that the alignment training stage proves crucial for establishing initial cross-lingual connections, while the representation finetuning stage refines these alignments for specific downstream tasks.

\subsection{Model Representation Visualization}
\label{sec:visualization}

\begin{figure}[t]
  \centering
  \subcaptionbox{E5-Mistral (language)}{\includegraphics[width=0.5\columnwidth]{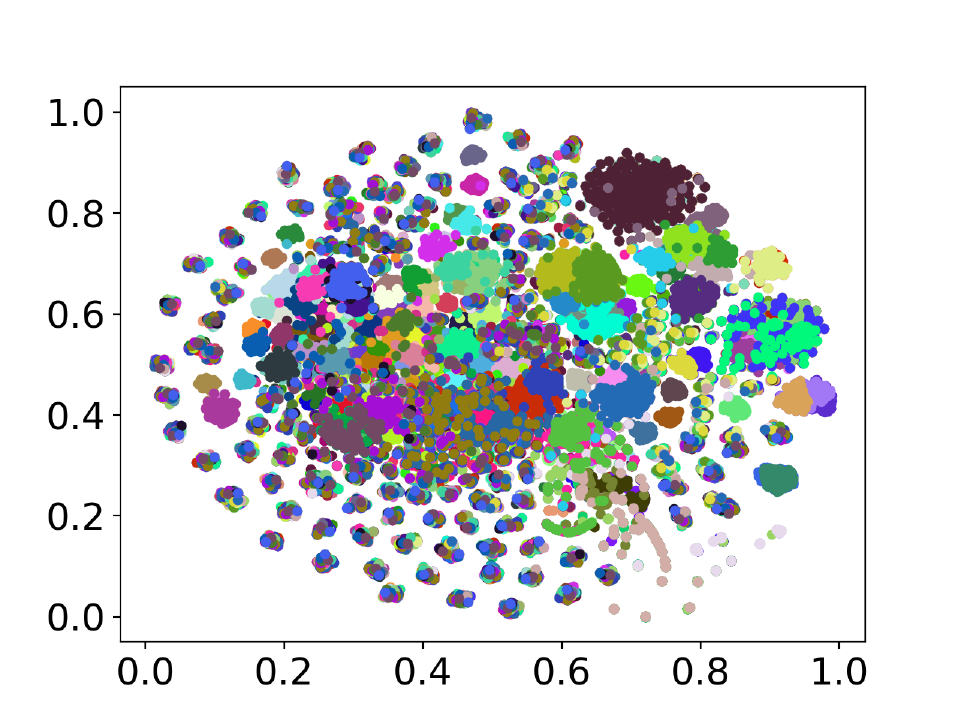}}%
  \hfill % <-- Seperation
  \subcaptionbox{LUSIFER (language)}{\includegraphics[width=0.5\columnwidth]{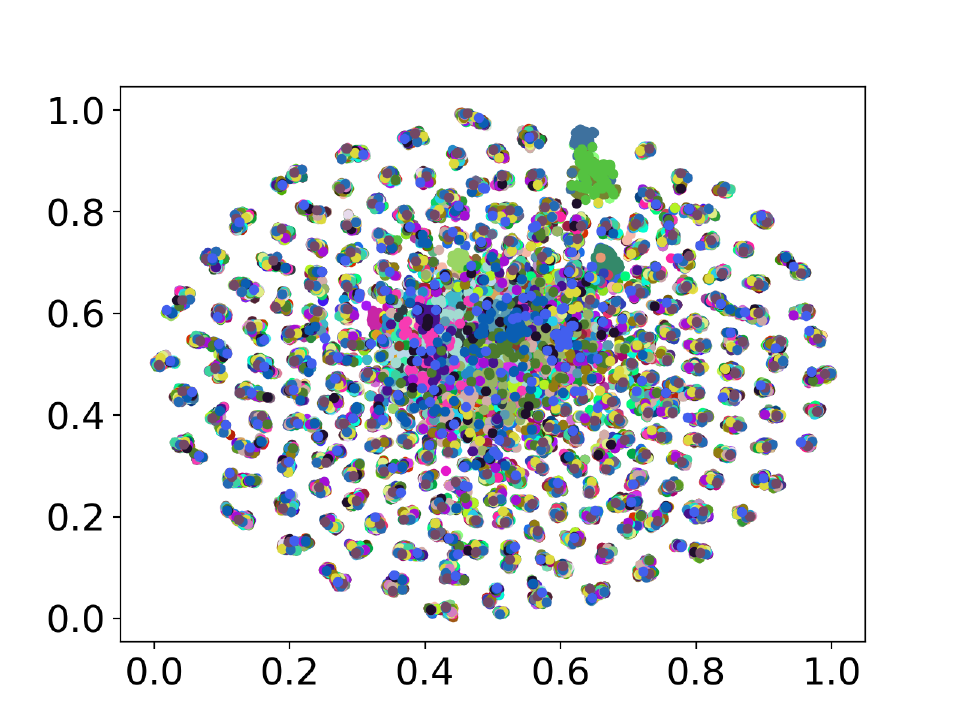}}
  \hfill
  % \subcaptionbox{E5-Mistral (label)}{\includegraphics[width=0.5\columnwidth]{tsne_sib200_e5_class.pdf}}%
  % \hfill % <-- Seperation
  % \subcaptionbox{LUSIFER (label)}{\includegraphics[width=0.5\columnwidth]{tsne_sib200_LUSIFER_class.pdf}}
  \caption{t-SNE representation of 200 randomly samples from the SIB200 dataset. The points are colored by the languages.}
  \label{fig:viz_tsne}
\end{figure}

%Figure \ref{fig:viz_tsne} presents 2D scatter plots of the representations from LUCIFER for 200 randomly sampled examples from the SIB200 dataset generated using t-SNE. 

Figure \ref{fig:viz_tsne} shows 2D scatter plots of representations from different models for 200 randomly sampled examples from the SIB200 dataset, visualized using t-SNE. The points are colored by the language of the samples. The t-SNE representation of E5-Mistral demonstrates a clearer separation between languages, with distinct clusters for each language. In contrast, the visualization of LUSIFER presents a more mixed distribution of languages, with overlapping clusters across different languages. This observation provides insights into LUSIFER's lingual-agnostic capabilities, highlighting the model's ability to bridge the gaps between representation spaces of different languages. These results suggest that LUSIFER's alignment strategy enables the model to comprehend semantics across multiple languages effectively, facilitating zero-shot multilingual transfer. Overall, our experiments confirm the advantages of the representation alignment strategies in LUSIFER in effectively enabling zero-shot multilingual transfer for LLM-based embedding methods. 

\section{Conclusion}
In this work, we propose LUSIFER, a novel framework that enables effective multilingual representation without explicit multilingual training data. LUSIFER aligns a multilingual encoder with a target English-centric LLM through a minimal set of trainable parameters, facilitating zero-shot multilingual transfer. Our experimental results demonstrate that LUSIFER achieves state-of-the-art performance across diverse languages and tasks, outperforming existing baseline models. Moreover, LUSIFER significantly enhances cross-lingual capabilities, enabling the model to process text-embedding tasks across multiple languages effectively. 

Our work provides a promising direction for enhancing multilingual representation capabilities in English-centric embedding models, enabling global applicability without requiring extensive multilingual training data. The alignment strategies employed in LUSIFER ensure that the model can comprehend and process semantic information across different languages, making it a versatile tool for various multilingual applications. Furthermore, the ability of LUSIFER to perform zero-shot multilingual transfer opens up new possibilities for natural language processing tasks in low-resource languages, where obtaining large-scale annotated data is often challenging. By leveraging the strengths of both multilingual encoders and English-centric LLMs, LUSIFER bridges the gap between languages, fostering better multilingual representation learning.

In future work, we plan to explore additional alignment strategies and further investigate the impact of LUSIFER's components on multilingual representation quality. We also aim to extend our framework to support different modalities such as images and audio and evaluate its performance on a wider range of tasks. Additionally, we will explore the integration of LUSIFER with other state-of-the-art models to further enhance its capabilities. Moreover, we anticipate that LUSIFER will facilitate broader applications of LLM embeddings in downstream tasks, ranging from deep context understanding requirements like sentiment analysis \cite{gupta2024comprehensivestudysentimentanalysis} to text style comprehension tasks such as authorship attribution \cite{rivera-soto-etal-2021-learning, 10.1145/3626772.3657956}, thereby contributing to the advancement of natural language processing and information retrieval fields.

\section*{Acknowledgements}

This research has been supported by the NSF grant \# 2239570. This research is also supported in part by the Office of the Director of National Intelligence (ODNI), Intelligence Advanced Research Projects Activity (IARPA), via the HIATUS Program contract 2022-22072200003. The views and conclusions contained herein are those of the authors and should not be interpreted as necessarily representing the official policies, either expressed or implied, of ODNI, IARPA, or the U.S. Government. The U.S. Government is authorized to reproduce and distribute reprints for governmental purposes, notwithstanding any copyright annotation therein.

\bibliographystyle{ACM-Reference-Format}
\balance
\bibliography{custom}


\begin{thebibliography}{74}

%%% the Web address.

\ifx \showCODEN    \undefined \def \showCODEN     #1{\unskip}     \fi
\ifx \showISBNx    \undefined \def \showISBNx     #1{\unskip}     \fi
\ifx \showISBNxiii \undefined \def \showISBNxiii  #1{\unskip}     \fi
\ifx \showISSN     \undefined \def \showISSN      #1{\unskip}     \fi
\ifx \showLCCN     \undefined \def \showLCCN      #1{\unskip}     \fi
\ifx \shownote     \undefined \def \shownote      #1{#1}          \fi
\ifx \showarticletitle \undefined \def \showarticletitle #1{#1}   \fi
\ifx \showURL      \undefined \def \showURL       {\relax}        \fi
% invisible to TeX
\providecommand\bibfield[2]{#2}
\providecommand\bibinfo[2]{#2}
\providecommand\natexlab[1]{#1}
\providecommand\showeprint[2][]{arXiv:#2}

\bibitem[Adelani et~al\mbox{.}(2024)]%
        {adelani2024sib200simpleinclusivebig}
\bibfield{author}{\bibinfo{person}{David~Ifeoluwa Adelani}, \bibinfo{person}{Hannah Liu}, \bibinfo{person}{Xiaoyu Shen}, \bibinfo{person}{Nikita Vassilyev}, \bibinfo{person}{Jesujoba~O. Alabi}, \bibinfo{person}{Yanke Mao}, \bibinfo{person}{Haonan Gao}, {and} \bibinfo{person}{Annie En-Shiun Lee}.} \bibinfo{year}{2024}\natexlab{}.
\newblock \bibinfo{title}{SIB-200: A Simple, Inclusive, and Big Evaluation Dataset for Topic Classification in 200+ Languages and Dialects}.
\newblock
\showeprint[arxiv]{2309.07445}~[cs.CL]
\urldef\tempurl%
\url{https://arxiv.org/abs/2309.07445}
\showURL{%
\tempurl}


\bibitem[Agirre et~al\mbox{.}(2016)]%
        {agirre-etal-2016-semeval}
\bibfield{author}{\bibinfo{person}{Eneko Agirre}, \bibinfo{person}{Carmen Banea}, \bibinfo{person}{Daniel Cer}, \bibinfo{person}{Mona Diab}, \bibinfo{person}{Aitor Gonzalez-Agirre}, \bibinfo{person}{Rada Mihalcea}, \bibinfo{person}{German Rigau}, {and} \bibinfo{person}{Janyce Wiebe}.} \bibinfo{year}{2016}\natexlab{}.
\newblock \showarticletitle{{S}em{E}val-2016 Task 1: Semantic Textual Similarity, Monolingual and Cross-Lingual Evaluation}. In \bibinfo{booktitle}{\emph{Proceedings of the 10th International Workshop on Semantic Evaluation ({S}em{E}val-2016)}}, \bibfield{editor}{\bibinfo{person}{Steven Bethard}, \bibinfo{person}{Marine Carpuat}, \bibinfo{person}{Daniel Cer}, \bibinfo{person}{David Jurgens}, \bibinfo{person}{Preslav Nakov}, {and} \bibinfo{person}{Torsten Zesch}} (Eds.). \bibinfo{publisher}{Association for Computational Linguistics}, \bibinfo{address}{San Diego, California}, \bibinfo{pages}{497--511}.
\newblock
\href{https://doi.org/10.18653/v1/S16-1081}{doi:\nolinkurl{10.18653/v1/S16-1081}}


\bibitem[Alayrac et~al\mbox{.}(2022)]%
        {NEURIPS2022_960a172b}
\bibfield{author}{\bibinfo{person}{Jean-Baptiste Alayrac}, \bibinfo{person}{Jeff Donahue}, \bibinfo{person}{Pauline Luc}, \bibinfo{person}{Antoine Miech}, \bibinfo{person}{Iain Barr}, \bibinfo{person}{Yana Hasson}, \bibinfo{person}{Karel Lenc}, \bibinfo{person}{Arthur Mensch}, \bibinfo{person}{Katherine Millican}, \bibinfo{person}{Malcolm Reynolds}, \bibinfo{person}{Roman Ring}, \bibinfo{person}{Eliza Rutherford}, \bibinfo{person}{Serkan Cabi}, \bibinfo{person}{Tengda Han}, \bibinfo{person}{Zhitao Gong}, \bibinfo{person}{Sina Samangooei}, \bibinfo{person}{Marianne Monteiro}, \bibinfo{person}{Jacob~L Menick}, \bibinfo{person}{Sebastian Borgeaud}, \bibinfo{person}{Andy Brock}, \bibinfo{person}{Aida Nematzadeh}, \bibinfo{person}{Sahand Sharifzadeh}, \bibinfo{person}{Miko\l~aj Bi\'{n}kowski}, \bibinfo{person}{Ricardo Barreira}, \bibinfo{person}{Oriol Vinyals}, \bibinfo{person}{Andrew Zisserman}, {and} \bibinfo{person}{Kar\'{e}n Simonyan}.} \bibinfo{year}{2022}\natexlab{}.
\newblock \showarticletitle{Flamingo: a Visual Language Model for Few-Shot Learning}. In \bibinfo{booktitle}{\emph{Advances in Neural Information Processing Systems}}, \bibfield{editor}{\bibinfo{person}{S.~Koyejo}, \bibinfo{person}{S.~Mohamed}, \bibinfo{person}{A.~Agarwal}, \bibinfo{person}{D.~Belgrave}, \bibinfo{person}{K.~Cho}, {and} \bibinfo{person}{A.~Oh}} (Eds.), Vol.~\bibinfo{volume}{35}. \bibinfo{publisher}{Curran Associates, Inc.}, \bibinfo{pages}{23716--23736}.
\newblock
\urldef\tempurl%
\url{https://proceedings.neurips.cc/paper_files/paper/2022/file/960a172bc7fbf0177ccccbb411a7d800-Paper-Conference.pdf}
\showURL{%
\tempurl}


\bibitem[Artetxe et~al\mbox{.}(2020)]%
        {artetxe-etal-2020-cross}
\bibfield{author}{\bibinfo{person}{Mikel Artetxe}, \bibinfo{person}{Sebastian Ruder}, {and} \bibinfo{person}{Dani Yogatama}.} \bibinfo{year}{2020}\natexlab{}.
\newblock \showarticletitle{On the Cross-lingual Transferability of Monolingual Representations}. In \bibinfo{booktitle}{\emph{Proceedings of the 58th Annual Meeting of the Association for Computational Linguistics}}, \bibfield{editor}{\bibinfo{person}{Dan Jurafsky}, \bibinfo{person}{Joyce Chai}, \bibinfo{person}{Natalie Schluter}, {and} \bibinfo{person}{Joel Tetreault}} (Eds.). \bibinfo{publisher}{Association for Computational Linguistics}, \bibinfo{address}{Online}, \bibinfo{pages}{4623--4637}.
\newblock
\href{https://doi.org/10.18653/v1/2020.acl-main.421}{doi:\nolinkurl{10.18653/v1/2020.acl-main.421}}


\bibitem[Bajaj et~al\mbox{.}(2018)]%
        {bajaj2018msmarcohumangenerated}
\bibfield{author}{\bibinfo{person}{Payal Bajaj}, \bibinfo{person}{Daniel Campos}, \bibinfo{person}{Nick Craswell}, \bibinfo{person}{Li Deng}, \bibinfo{person}{Jianfeng Gao}, \bibinfo{person}{Xiaodong Liu}, \bibinfo{person}{Rangan Majumder}, \bibinfo{person}{Andrew McNamara}, \bibinfo{person}{Bhaskar Mitra}, \bibinfo{person}{Tri Nguyen}, \bibinfo{person}{Mir Rosenberg}, \bibinfo{person}{Xia Song}, \bibinfo{person}{Alina Stoica}, \bibinfo{person}{Saurabh Tiwary}, {and} \bibinfo{person}{Tong Wang}.} \bibinfo{year}{2018}\natexlab{}.
\newblock \bibinfo{title}{MS MARCO: A Human Generated MAchine Reading COmprehension Dataset}.
\newblock
\showeprint[arxiv]{1611.09268}~[cs.CL]
\urldef\tempurl%
\url{https://arxiv.org/abs/1611.09268}
\showURL{%
\tempurl}


\bibitem[Bandarkar et~al\mbox{.}(2024)]%
        {Bandarkar2024}
\bibfield{author}{\bibinfo{person}{Lucas Bandarkar}, \bibinfo{person}{Davis Liang}, \bibinfo{person}{Benjamin Muller}, \bibinfo{person}{Mikel Artetxe}, \bibinfo{person}{Satya~Narayan Shukla}, \bibinfo{person}{Donald Husa}, \bibinfo{person}{Naman Goyal}, \bibinfo{person}{Abhinandan Krishnan}, \bibinfo{person}{Luke Zettlemoyer}, {and} \bibinfo{person}{Madian Khabsa}.} \bibinfo{year}{2024}\natexlab{}.
\newblock \showarticletitle{The Belebele Benchmark: a Parallel Reading Comprehension Dataset in 122 Language Variants}. In \bibinfo{booktitle}{\emph{Proceedings of the 62nd Annual Meeting of the Association for Computational Linguistics (Volume 1: Long Papers)}}. \bibinfo{publisher}{Association for Computational Linguistics}, \bibinfo{pages}{749–775}.
\newblock
\href{https://doi.org/10.18653/v1/2024.acl-long.44}{doi:\nolinkurl{10.18653/v1/2024.acl-long.44}}


\bibitem[Bansal et~al\mbox{.}(2024)]%
        {bansal2024llm}
\bibfield{author}{\bibinfo{person}{Rachit Bansal}, \bibinfo{person}{Bidisha Samanta}, \bibinfo{person}{Siddharth Dalmia}, \bibinfo{person}{Nitish Gupta}, \bibinfo{person}{Sriram Ganapathy}, \bibinfo{person}{Abhishek Bapna}, \bibinfo{person}{Prateek Jain}, {and} \bibinfo{person}{Partha Talukdar}.} \bibinfo{year}{2024}\natexlab{}.
\newblock \showarticletitle{{LLM} Augmented {LLM}s: Expanding Capabilities through Composition}. In \bibinfo{booktitle}{\emph{The Twelfth International Conference on Learning Representations}}.
\newblock
\urldef\tempurl%
\url{https://openreview.net/forum?id=jjA4O1vJRz}
\showURL{%
\tempurl}


\bibitem[BehnamGhader et~al\mbox{.}(2024)]%
        {behnamghader2024llm2veclargelanguagemodels}
\bibfield{author}{\bibinfo{person}{Parishad BehnamGhader}, \bibinfo{person}{Vaibhav Adlakha}, \bibinfo{person}{Marius Mosbach}, \bibinfo{person}{Dzmitry Bahdanau}, \bibinfo{person}{Nicolas Chapados}, {and} \bibinfo{person}{Siva Reddy}.} \bibinfo{year}{2024}\natexlab{}.
\newblock \bibinfo{title}{LLM2Vec: Large Language Models Are Secretly Powerful Text Encoders}.
\newblock
\showeprint[arxiv]{2404.05961}~[cs.CL]
\urldef\tempurl%
\url{https://arxiv.org/abs/2404.05961}
\showURL{%
\tempurl}


\bibitem[Bowman et~al\mbox{.}(2015)]%
        {bowman-etal-2015-large}
\bibfield{author}{\bibinfo{person}{Samuel~R. Bowman}, \bibinfo{person}{Gabor Angeli}, \bibinfo{person}{Christopher Potts}, {and} \bibinfo{person}{Christopher~D. Manning}.} \bibinfo{year}{2015}\natexlab{}.
\newblock \showarticletitle{A large annotated corpus for learning natural language inference}. In \bibinfo{booktitle}{\emph{Proceedings of the 2015 Conference on Empirical Methods in Natural Language Processing}}, \bibfield{editor}{\bibinfo{person}{Llu{\'\i}s M{\`a}rquez}, \bibinfo{person}{Chris Callison-Burch}, {and} \bibinfo{person}{Jian Su}} (Eds.). \bibinfo{publisher}{Association for Computational Linguistics}, \bibinfo{address}{Lisbon, Portugal}, \bibinfo{pages}{632--642}.
\newblock
\href{https://doi.org/10.18653/v1/D15-1075}{doi:\nolinkurl{10.18653/v1/D15-1075}}


\bibitem[Chen et~al\mbox{.}(2024)]%
        {bge-m3}
\bibfield{author}{\bibinfo{person}{Jianlv Chen}, \bibinfo{person}{Shitao Xiao}, \bibinfo{person}{Peitian Zhang}, \bibinfo{person}{Kun Luo}, \bibinfo{person}{Defu Lian}, {and} \bibinfo{person}{Zheng Liu}.} \bibinfo{year}{2024}\natexlab{}.
\newblock \bibinfo{title}{BGE M3-Embedding: Multi-Lingual, Multi-Functionality, Multi-Granularity Text Embeddings Through Self-Knowledge Distillation}.
\newblock
\showeprint[arxiv]{2402.03216}~[cs.CL]


\bibitem[Conneau et~al\mbox{.}(2020)]%
        {conneau-etal-2020-unsupervised}
\bibfield{author}{\bibinfo{person}{Alexis Conneau}, \bibinfo{person}{Kartikay Khandelwal}, \bibinfo{person}{Naman Goyal}, \bibinfo{person}{Vishrav Chaudhary}, \bibinfo{person}{Guillaume Wenzek}, \bibinfo{person}{Francisco Guzm{\'a}n}, \bibinfo{person}{Edouard Grave}, \bibinfo{person}{Myle Ott}, \bibinfo{person}{Luke Zettlemoyer}, {and} \bibinfo{person}{Veselin Stoyanov}.} \bibinfo{year}{2020}\natexlab{}.
\newblock \showarticletitle{Unsupervised Cross-lingual Representation Learning at Scale}. In \bibinfo{booktitle}{\emph{Proceedings of the 58th Annual Meeting of the Association for Computational Linguistics}}, \bibfield{editor}{\bibinfo{person}{Dan Jurafsky}, \bibinfo{person}{Joyce Chai}, \bibinfo{person}{Natalie Schluter}, {and} \bibinfo{person}{Joel Tetreault}} (Eds.). \bibinfo{publisher}{Association for Computational Linguistics}, \bibinfo{address}{Online}, \bibinfo{pages}{8440--8451}.
\newblock
\href{https://doi.org/10.18653/v1/2020.acl-main.747}{doi:\nolinkurl{10.18653/v1/2020.acl-main.747}}


\bibitem[Conneau et~al\mbox{.}(2018)]%
        {conneau-etal-2018-xnli}
\bibfield{author}{\bibinfo{person}{Alexis Conneau}, \bibinfo{person}{Ruty Rinott}, \bibinfo{person}{Guillaume Lample}, \bibinfo{person}{Adina Williams}, \bibinfo{person}{Samuel Bowman}, \bibinfo{person}{Holger Schwenk}, {and} \bibinfo{person}{Veselin Stoyanov}.} \bibinfo{year}{2018}\natexlab{}.
\newblock \showarticletitle{{XNLI}: Evaluating Cross-lingual Sentence Representations}. In \bibinfo{booktitle}{\emph{Proceedings of the 2018 Conference on Empirical Methods in Natural Language Processing}}, \bibfield{editor}{\bibinfo{person}{Ellen Riloff}, \bibinfo{person}{David Chiang}, \bibinfo{person}{Julia Hockenmaier}, {and} \bibinfo{person}{Jun{'}ichi Tsujii}} (Eds.). \bibinfo{publisher}{Association for Computational Linguistics}, \bibinfo{address}{Brussels, Belgium}, \bibinfo{pages}{2475--2485}.
\newblock
\href{https://doi.org/10.18653/v1/D18-1269}{doi:\nolinkurl{10.18653/v1/D18-1269}}


\bibitem[Devlin et~al\mbox{.}(2019)]%
        {devlin-etal-2019-bert}
\bibfield{author}{\bibinfo{person}{Jacob Devlin}, \bibinfo{person}{Ming-Wei Chang}, \bibinfo{person}{Kenton Lee}, {and} \bibinfo{person}{Kristina Toutanova}.} \bibinfo{year}{2019}\natexlab{}.
\newblock \showarticletitle{{BERT}: Pre-training of Deep Bidirectional Transformers for Language Understanding}. In \bibinfo{booktitle}{\emph{Proceedings of the 2019 Conference of the North {A}merican Chapter of the Association for Computational Linguistics: Human Language Technologies, Volume 1 (Long and Short Papers)}}, \bibfield{editor}{\bibinfo{person}{Jill Burstein}, \bibinfo{person}{Christy Doran}, {and} \bibinfo{person}{Thamar Solorio}} (Eds.). \bibinfo{publisher}{Association for Computational Linguistics}, \bibinfo{address}{Minneapolis, Minnesota}, \bibinfo{pages}{4171--4186}.
\newblock
\href{https://doi.org/10.18653/v1/N19-1423}{doi:\nolinkurl{10.18653/v1/N19-1423}}


\bibitem[Falcon and team(2024)]%
        {falcon_2024_13254264}
\bibfield{author}{\bibinfo{person}{William Falcon} {and} \bibinfo{person}{The PyTorch~Lightning team}.} \bibinfo{year}{2024}\natexlab{}.
\newblock \bibinfo{booktitle}{\emph{PyTorch Lightning}}.
\newblock
\href{https://doi.org/10.5281/zenodo.13254264}{doi:\nolinkurl{10.5281/zenodo.13254264}}


\bibitem[Gao et~al\mbox{.}(2021b)]%
        {gao-etal-2021-scaling}
\bibfield{author}{\bibinfo{person}{Luyu Gao}, \bibinfo{person}{Yunyi Zhang}, \bibinfo{person}{Jiawei Han}, {and} \bibinfo{person}{Jamie Callan}.} \bibinfo{year}{2021}\natexlab{b}.
\newblock \showarticletitle{Scaling Deep Contrastive Learning Batch Size under Memory Limited Setup}. In \bibinfo{booktitle}{\emph{Proceedings of the 6th Workshop on Representation Learning for NLP (RepL4NLP-2021)}}, \bibfield{editor}{\bibinfo{person}{Anna Rogers}, \bibinfo{person}{Iacer Calixto}, \bibinfo{person}{Ivan Vuli{\'c}}, \bibinfo{person}{Naomi Saphra}, \bibinfo{person}{Nora Kassner}, \bibinfo{person}{Oana-Maria Camburu}, \bibinfo{person}{Trapit Bansal}, {and} \bibinfo{person}{Vered Shwartz}} (Eds.). \bibinfo{publisher}{Association for Computational Linguistics}, \bibinfo{address}{Online}, \bibinfo{pages}{316--321}.
\newblock
\href{https://doi.org/10.18653/v1/2021.repl4nlp-1.31}{doi:\nolinkurl{10.18653/v1/2021.repl4nlp-1.31}}


\bibitem[Gao et~al\mbox{.}(2021a)]%
        {gao-etal-2021-simcse}
\bibfield{author}{\bibinfo{person}{Tianyu Gao}, \bibinfo{person}{Xingcheng Yao}, {and} \bibinfo{person}{Danqi Chen}.} \bibinfo{year}{2021}\natexlab{a}.
\newblock \showarticletitle{{S}im{CSE}: Simple Contrastive Learning of Sentence Embeddings}. In \bibinfo{booktitle}{\emph{Proceedings of the 2021 Conference on Empirical Methods in Natural Language Processing}}, \bibfield{editor}{\bibinfo{person}{Marie-Francine Moens}, \bibinfo{person}{Xuanjing Huang}, \bibinfo{person}{Lucia Specia}, {and} \bibinfo{person}{Scott Wen-tau Yih}} (Eds.). \bibinfo{publisher}{Association for Computational Linguistics}, \bibinfo{address}{Online and Punta Cana, Dominican Republic}, \bibinfo{pages}{6894--6910}.
\newblock
\href{https://doi.org/10.18653/v1/2021.emnlp-main.552}{doi:\nolinkurl{10.18653/v1/2021.emnlp-main.552}}


\bibitem[Gao et~al\mbox{.}(2024)]%
        {gao2024retrievalaugmentedgenerationlargelanguage}
\bibfield{author}{\bibinfo{person}{Yunfan Gao}, \bibinfo{person}{Yun Xiong}, \bibinfo{person}{Xinyu Gao}, \bibinfo{person}{Kangxiang Jia}, \bibinfo{person}{Jinliu Pan}, \bibinfo{person}{Yuxi Bi}, \bibinfo{person}{Yi Dai}, \bibinfo{person}{Jiawei Sun}, \bibinfo{person}{Meng Wang}, {and} \bibinfo{person}{Haofen Wang}.} \bibinfo{year}{2024}\natexlab{}.
\newblock \bibinfo{title}{Retrieval-Augmented Generation for Large Language Models: A Survey}.
\newblock
\showeprint[arxiv]{2312.10997}~[cs.CL]
\urldef\tempurl%
\url{https://arxiv.org/abs/2312.10997}
\showURL{%
\tempurl}


\bibitem[Gupta et~al\mbox{.}(2024)]%
        {gupta2024comprehensivestudysentimentanalysis}
\bibfield{author}{\bibinfo{person}{Shailja Gupta}, \bibinfo{person}{Rajesh Ranjan}, {and} \bibinfo{person}{Surya~Narayan Singh}.} \bibinfo{year}{2024}\natexlab{}.
\newblock \bibinfo{title}{Comprehensive Study on Sentiment Analysis: From Rule-based to modern LLM based system}.
\newblock
\showeprint[arxiv]{2409.09989}~[cs.CL]
\urldef\tempurl%
\url{https://arxiv.org/abs/2409.09989}
\showURL{%
\tempurl}


\bibitem[Hu et~al\mbox{.}(2022)]%
        {hu2022lora}
\bibfield{author}{\bibinfo{person}{Edward~J Hu}, \bibinfo{person}{yelong shen}, \bibinfo{person}{Phillip Wallis}, \bibinfo{person}{Zeyuan Allen-Zhu}, \bibinfo{person}{Yuanzhi Li}, \bibinfo{person}{Shean Wang}, \bibinfo{person}{Lu Wang}, {and} \bibinfo{person}{Weizhu Chen}.} \bibinfo{year}{2022}\natexlab{}.
\newblock \showarticletitle{Lo{RA}: Low-Rank Adaptation of Large Language Models}. In \bibinfo{booktitle}{\emph{International Conference on Learning Representations}}.
\newblock
\urldef\tempurl%
\url{https://openreview.net/forum?id=nZeVKeeFYf9}
\showURL{%
\tempurl}


\bibitem[Izacard et~al\mbox{.}(2022)]%
        {izacard2022unsuperviseddenseinformationretrieval}
\bibfield{author}{\bibinfo{person}{Gautier Izacard}, \bibinfo{person}{Mathilde Caron}, \bibinfo{person}{Lucas Hosseini}, \bibinfo{person}{Sebastian Riedel}, \bibinfo{person}{Piotr Bojanowski}, \bibinfo{person}{Armand Joulin}, {and} \bibinfo{person}{Edouard Grave}.} \bibinfo{year}{2022}\natexlab{}.
\newblock \bibinfo{title}{Unsupervised Dense Information Retrieval with Contrastive Learning}.
\newblock
\showeprint[arxiv]{2112.09118}~[cs.IR]
\urldef\tempurl%
\url{https://arxiv.org/abs/2112.09118}
\showURL{%
\tempurl}


\bibitem[Jiang et~al\mbox{.}(2023)]%
        {jiang2023mistral7b}
\bibfield{author}{\bibinfo{person}{Albert~Q. Jiang}, \bibinfo{person}{Alexandre Sablayrolles}, \bibinfo{person}{Arthur Mensch}, \bibinfo{person}{Chris Bamford}, \bibinfo{person}{Devendra~Singh Chaplot}, \bibinfo{person}{Diego de~las Casas}, \bibinfo{person}{Florian Bressand}, \bibinfo{person}{Gianna Lengyel}, \bibinfo{person}{Guillaume Lample}, \bibinfo{person}{Lucile Saulnier}, \bibinfo{person}{Lélio~Renard Lavaud}, \bibinfo{person}{Marie-Anne Lachaux}, \bibinfo{person}{Pierre Stock}, \bibinfo{person}{Teven~Le Scao}, \bibinfo{person}{Thibaut Lavril}, \bibinfo{person}{Thomas Wang}, \bibinfo{person}{Timothée Lacroix}, {and} \bibinfo{person}{William~El Sayed}.} \bibinfo{year}{2023}\natexlab{}.
\newblock \bibinfo{title}{Mistral 7B}.
\newblock
\showeprint[arxiv]{2310.06825}~[cs.CL]
\urldef\tempurl%
\url{https://arxiv.org/abs/2310.06825}
\showURL{%
\tempurl}


\bibitem[Karpukhin et~al\mbox{.}(2020)]%
        {karpukhin-etal-2020-dense}
\bibfield{author}{\bibinfo{person}{Vladimir Karpukhin}, \bibinfo{person}{Barlas Oguz}, \bibinfo{person}{Sewon Min}, \bibinfo{person}{Patrick Lewis}, \bibinfo{person}{Ledell Wu}, \bibinfo{person}{Sergey Edunov}, \bibinfo{person}{Danqi Chen}, {and} \bibinfo{person}{Wen-tau Yih}.} \bibinfo{year}{2020}\natexlab{}.
\newblock \showarticletitle{Dense Passage Retrieval for Open-Domain Question Answering}. In \bibinfo{booktitle}{\emph{Proceedings of the 2020 Conference on Empirical Methods in Natural Language Processing (EMNLP)}}, \bibfield{editor}{\bibinfo{person}{Bonnie Webber}, \bibinfo{person}{Trevor Cohn}, \bibinfo{person}{Yulan He}, {and} \bibinfo{person}{Yang Liu}} (Eds.). \bibinfo{publisher}{Association for Computational Linguistics}, \bibinfo{address}{Online}, \bibinfo{pages}{6769--6781}.
\newblock
\href{https://doi.org/10.18653/v1/2020.emnlp-main.550}{doi:\nolinkurl{10.18653/v1/2020.emnlp-main.550}}


\bibitem[Kwiatkowski et~al\mbox{.}(2019)]%
        {kwiatkowski-etal-2019-natural}
\bibfield{author}{\bibinfo{person}{Tom Kwiatkowski}, \bibinfo{person}{Jennimaria Palomaki}, \bibinfo{person}{Olivia Redfield}, \bibinfo{person}{Michael Collins}, \bibinfo{person}{Ankur Parikh}, \bibinfo{person}{Chris Alberti}, \bibinfo{person}{Danielle Epstein}, \bibinfo{person}{Illia Polosukhin}, \bibinfo{person}{Jacob Devlin}, \bibinfo{person}{Kenton Lee}, \bibinfo{person}{Kristina Toutanova}, \bibinfo{person}{Llion Jones}, \bibinfo{person}{Matthew Kelcey}, \bibinfo{person}{Ming-Wei Chang}, \bibinfo{person}{Andrew~M. Dai}, \bibinfo{person}{Jakob Uszkoreit}, \bibinfo{person}{Quoc Le}, {and} \bibinfo{person}{Slav Petrov}.} \bibinfo{year}{2019}\natexlab{}.
\newblock \showarticletitle{Natural Questions: A Benchmark for Question Answering Research}.
\newblock \bibinfo{journal}{\emph{Transactions of the Association for Computational Linguistics}}  \bibinfo{volume}{7} (\bibinfo{year}{2019}), \bibinfo{pages}{452--466}.
\newblock
\href{https://doi.org/10.1162/tacl_a_00276}{doi:\nolinkurl{10.1162/tacl_a_00276}}


\bibitem[Lai et~al\mbox{.}(2023)]%
        {lai-etal-2023-chatgpt}
\bibfield{author}{\bibinfo{person}{Viet~Dac Lai}, \bibinfo{person}{Nghia Ngo}, \bibinfo{person}{Amir Pouran Ben~Veyseh}, \bibinfo{person}{Hieu Man}, \bibinfo{person}{Franck Dernoncourt}, \bibinfo{person}{Trung Bui}, {and} \bibinfo{person}{Thien~Huu Nguyen}.} \bibinfo{year}{2023}\natexlab{}.
\newblock \showarticletitle{{C}hat{GPT} Beyond {E}nglish: Towards a Comprehensive Evaluation of Large Language Models in Multilingual Learning}. In \bibinfo{booktitle}{\emph{Findings of the Association for Computational Linguistics: EMNLP 2023}}, \bibfield{editor}{\bibinfo{person}{Houda Bouamor}, \bibinfo{person}{Juan Pino}, {and} \bibinfo{person}{Kalika Bali}} (Eds.). \bibinfo{publisher}{Association for Computational Linguistics}, \bibinfo{address}{Singapore}, \bibinfo{pages}{13171--13189}.
\newblock
\href{https://doi.org/10.18653/v1/2023.findings-emnlp.878}{doi:\nolinkurl{10.18653/v1/2023.findings-emnlp.878}}


\bibitem[Lee et~al\mbox{.}(2024)]%
        {lee2024nvembedimprovedtechniquestraining}
\bibfield{author}{\bibinfo{person}{Chankyu Lee}, \bibinfo{person}{Rajarshi Roy}, \bibinfo{person}{Mengyao Xu}, \bibinfo{person}{Jonathan Raiman}, \bibinfo{person}{Mohammad Shoeybi}, \bibinfo{person}{Bryan Catanzaro}, {and} \bibinfo{person}{Wei Ping}.} \bibinfo{year}{2024}\natexlab{}.
\newblock \bibinfo{title}{NV-Embed: Improved Techniques for Training LLMs as Generalist Embedding Models}.
\newblock
\showeprint[arxiv]{2405.17428}~[cs.CL]
\urldef\tempurl%
\url{https://arxiv.org/abs/2405.17428}
\showURL{%
\tempurl}


\bibitem[Lewis et~al\mbox{.}(2020a)]%
        {lewis-etal-2020-mlqa}
\bibfield{author}{\bibinfo{person}{Patrick Lewis}, \bibinfo{person}{Barlas Oguz}, \bibinfo{person}{Ruty Rinott}, \bibinfo{person}{Sebastian Riedel}, {and} \bibinfo{person}{Holger Schwenk}.} \bibinfo{year}{2020}\natexlab{a}.
\newblock \showarticletitle{{MLQA}: Evaluating Cross-lingual Extractive Question Answering}. In \bibinfo{booktitle}{\emph{Proceedings of the 58th Annual Meeting of the Association for Computational Linguistics}}, \bibfield{editor}{\bibinfo{person}{Dan Jurafsky}, \bibinfo{person}{Joyce Chai}, \bibinfo{person}{Natalie Schluter}, {and} \bibinfo{person}{Joel Tetreault}} (Eds.). \bibinfo{publisher}{Association for Computational Linguistics}, \bibinfo{address}{Online}, \bibinfo{pages}{7315--7330}.
\newblock
\href{https://doi.org/10.18653/v1/2020.acl-main.653}{doi:\nolinkurl{10.18653/v1/2020.acl-main.653}}


\bibitem[Lewis et~al\mbox{.}(2020b)]%
        {NEURIPS2020_6b493230}
\bibfield{author}{\bibinfo{person}{Patrick Lewis}, \bibinfo{person}{Ethan Perez}, \bibinfo{person}{Aleksandra Piktus}, \bibinfo{person}{Fabio Petroni}, \bibinfo{person}{Vladimir Karpukhin}, \bibinfo{person}{Naman Goyal}, \bibinfo{person}{Heinrich K\"{u}ttler}, \bibinfo{person}{Mike Lewis}, \bibinfo{person}{Wen-tau Yih}, \bibinfo{person}{Tim Rockt\"{a}schel}, \bibinfo{person}{Sebastian Riedel}, {and} \bibinfo{person}{Douwe Kiela}.} \bibinfo{year}{2020}\natexlab{b}.
\newblock \showarticletitle{Retrieval-Augmented Generation for Knowledge-Intensive NLP Tasks}. In \bibinfo{booktitle}{\emph{Advances in Neural Information Processing Systems}}, \bibfield{editor}{\bibinfo{person}{H.~Larochelle}, \bibinfo{person}{M.~Ranzato}, \bibinfo{person}{R.~Hadsell}, \bibinfo{person}{M.F. Balcan}, {and} \bibinfo{person}{H.~Lin}} (Eds.), Vol.~\bibinfo{volume}{33}. \bibinfo{publisher}{Curran Associates, Inc.}, \bibinfo{pages}{9459--9474}.
\newblock
\urldef\tempurl%
\url{https://proceedings.neurips.cc/paper_files/paper/2020/file/6b493230205f780e1bc26945df7481e5-Paper.pdf}
\showURL{%
\tempurl}


\bibitem[Lewis et~al\mbox{.}(2021)]%
        {lewis-etal-2021-paq}
\bibfield{author}{\bibinfo{person}{Patrick Lewis}, \bibinfo{person}{Yuxiang Wu}, \bibinfo{person}{Linqing Liu}, \bibinfo{person}{Pasquale Minervini}, \bibinfo{person}{Heinrich K{\"u}ttler}, \bibinfo{person}{Aleksandra Piktus}, \bibinfo{person}{Pontus Stenetorp}, {and} \bibinfo{person}{Sebastian Riedel}.} \bibinfo{year}{2021}\natexlab{}.
\newblock \showarticletitle{{PAQ}: 65 Million Probably-Asked Questions and What You Can Do With Them}.
\newblock \bibinfo{journal}{\emph{Transactions of the Association for Computational Linguistics}}  \bibinfo{volume}{9} (\bibinfo{year}{2021}), \bibinfo{pages}{1098--1115}.
\newblock
\href{https://doi.org/10.1162/tacl_a_00415}{doi:\nolinkurl{10.1162/tacl_a_00415}}


\bibitem[Li et~al\mbox{.}(2023)]%
        {li2023generaltextembeddingsmultistage}
\bibfield{author}{\bibinfo{person}{Zehan Li}, \bibinfo{person}{Xin Zhang}, \bibinfo{person}{Yanzhao Zhang}, \bibinfo{person}{Dingkun Long}, \bibinfo{person}{Pengjun Xie}, {and} \bibinfo{person}{Meishan Zhang}.} \bibinfo{year}{2023}\natexlab{}.
\newblock \bibinfo{title}{Towards General Text Embeddings with Multi-stage Contrastive Learning}.
\newblock
\showeprint[arxiv]{2308.03281}~[cs.CL]
\urldef\tempurl%
\url{https://arxiv.org/abs/2308.03281}
\showURL{%
\tempurl}


\bibitem[Libovick{\'y} et~al\mbox{.}(2020)]%
        {libovicky-etal-2020-language}
\bibfield{author}{\bibinfo{person}{Jind{\v{r}}ich Libovick{\'y}}, \bibinfo{person}{Rudolf Rosa}, {and} \bibinfo{person}{Alexander Fraser}.} \bibinfo{year}{2020}\natexlab{}.
\newblock \showarticletitle{On the Language Neutrality of Pre-trained Multilingual Representations}. In \bibinfo{booktitle}{\emph{Findings of the Association for Computational Linguistics: EMNLP 2020}}, \bibfield{editor}{\bibinfo{person}{Trevor Cohn}, \bibinfo{person}{Yulan He}, {and} \bibinfo{person}{Yang Liu}} (Eds.). \bibinfo{publisher}{Association for Computational Linguistics}, \bibinfo{address}{Online}, \bibinfo{pages}{1663--1674}.
\newblock
\href{https://doi.org/10.18653/v1/2020.findings-emnlp.150}{doi:\nolinkurl{10.18653/v1/2020.findings-emnlp.150}}


\bibitem[Liu et~al\mbox{.}(2024a)]%
        {liu2024visual}
\bibfield{author}{\bibinfo{person}{Haotian Liu}, \bibinfo{person}{Chunyuan Li}, \bibinfo{person}{Qingyang Wu}, {and} \bibinfo{person}{Yong~Jae Lee}.} \bibinfo{year}{2024}\natexlab{a}.
\newblock \showarticletitle{Visual instruction tuning}.
\newblock \bibinfo{journal}{\emph{Advances in neural information processing systems}}  \bibinfo{volume}{36} (\bibinfo{year}{2024}).
\newblock


\bibitem[Liu et~al\mbox{.}(2020)]%
        {liu-etal-2020-cross-lingual}
\bibfield{author}{\bibinfo{person}{Jiapeng Liu}, \bibinfo{person}{Xiao Zhang}, \bibinfo{person}{Dan Goldwasser}, {and} \bibinfo{person}{Xiao Wang}.} \bibinfo{year}{2020}\natexlab{}.
\newblock \showarticletitle{Cross-Lingual Document Retrieval with Smooth Learning}. In \bibinfo{booktitle}{\emph{Proceedings of the 28th International Conference on Computational Linguistics}}, \bibfield{editor}{\bibinfo{person}{Donia Scott}, \bibinfo{person}{Nuria Bel}, {and} \bibinfo{person}{Chengqing Zong}} (Eds.). \bibinfo{publisher}{International Committee on Computational Linguistics}, \bibinfo{address}{Barcelona, Spain (Online)}, \bibinfo{pages}{3616--3629}.
\newblock
\href{https://doi.org/10.18653/v1/2020.coling-main.323}{doi:\nolinkurl{10.18653/v1/2020.coling-main.323}}


\bibitem[Liu et~al\mbox{.}(2024b)]%
        {liu2024chatqa}
\bibfield{author}{\bibinfo{person}{Zihan Liu}, \bibinfo{person}{Wei Ping}, \bibinfo{person}{Rajarshi Roy}, \bibinfo{person}{Peng Xu}, \bibinfo{person}{Chankyu Lee}, \bibinfo{person}{Mohammad Shoeybi}, {and} \bibinfo{person}{Bryan Catanzaro}.} \bibinfo{year}{2024}\natexlab{b}.
\newblock \showarticletitle{Chatqa: Surpassing gpt-4 on conversational qa and rag}.
\newblock \bibinfo{journal}{\emph{arXiv preprint arXiv:2401.10225}} (\bibinfo{year}{2024}).
\newblock


\bibitem[Lu et~al\mbox{.}(2024)]%
        {lu2024ovisstructuralembeddingalignment}
\bibfield{author}{\bibinfo{person}{Shiyin Lu}, \bibinfo{person}{Yang Li}, \bibinfo{person}{Qing-Guo Chen}, \bibinfo{person}{Zhao Xu}, \bibinfo{person}{Weihua Luo}, \bibinfo{person}{Kaifu Zhang}, {and} \bibinfo{person}{Han-Jia Ye}.} \bibinfo{year}{2024}\natexlab{}.
\newblock \bibinfo{title}{Ovis: Structural Embedding Alignment for Multimodal Large Language Model}.
\newblock
\showeprint[arxiv]{2405.20797}~[cs.CV]
\urldef\tempurl%
\url{https://arxiv.org/abs/2405.20797}
\showURL{%
\tempurl}


\bibitem[Luo et~al\mbox{.}(2024)]%
        {luo2024largelanguagemodelsfoundations}
\bibfield{author}{\bibinfo{person}{Kun Luo}, \bibinfo{person}{Minghao Qin}, \bibinfo{person}{Zheng Liu}, \bibinfo{person}{Shitao Xiao}, \bibinfo{person}{Jun Zhao}, {and} \bibinfo{person}{Kang Liu}.} \bibinfo{year}{2024}\natexlab{}.
\newblock \bibinfo{title}{Large Language Models as Foundations for Next-Gen Dense Retrieval: A Comprehensive Empirical Assessment}.
\newblock
\showeprint[arxiv]{2408.12194}~[cs.CL]
\urldef\tempurl%
\url{https://arxiv.org/abs/2408.12194}
\showURL{%
\tempurl}


\bibitem[Maia et~al\mbox{.}(2018)]%
        {10.1145/3184558.3192301}
\bibfield{author}{\bibinfo{person}{Macedo Maia}, \bibinfo{person}{Siegfried Handschuh}, \bibinfo{person}{Andr\'{e} Freitas}, \bibinfo{person}{Brian Davis}, \bibinfo{person}{Ross McDermott}, \bibinfo{person}{Manel Zarrouk}, {and} \bibinfo{person}{Alexandra Balahur}.} \bibinfo{year}{2018}\natexlab{}.
\newblock \showarticletitle{WWW'18 Open Challenge: Financial Opinion Mining and Question Answering}. In \bibinfo{booktitle}{\emph{Companion Proceedings of the The Web Conference 2018}} (Lyon, France) \emph{(\bibinfo{series}{WWW '18})}. \bibinfo{publisher}{International World Wide Web Conferences Steering Committee}, \bibinfo{address}{Republic and Canton of Geneva, CHE}, \bibinfo{pages}{1941–1942}.
\newblock
\showISBNx{9781450356404}
\href{https://doi.org/10.1145/3184558.3192301}{doi:\nolinkurl{10.1145/3184558.3192301}}


\bibitem[Man et~al\mbox{.}(2024)]%
        {man2024ullmeunifiedframeworklarge}
\bibfield{author}{\bibinfo{person}{Hieu Man}, \bibinfo{person}{Nghia~Trung Ngo}, \bibinfo{person}{Franck Dernoncourt}, {and} \bibinfo{person}{Thien~Huu Nguyen}.} \bibinfo{year}{2024}\natexlab{}.
\newblock \bibinfo{title}{ULLME: A Unified Framework for Large Language Model Embeddings with Generation-Augmented Learning}.
\newblock
\showeprint[arxiv]{2408.03402}~[cs.CL]
\urldef\tempurl%
\url{https://arxiv.org/abs/2408.03402}
\showURL{%
\tempurl}


\bibitem[Man and Nguyen(2024)]%
        {10.1145/3626772.3657956}
\bibfield{author}{\bibinfo{person}{Hieu Man} {and} \bibinfo{person}{Thien~Huu Nguyen}.} \bibinfo{year}{2024}\natexlab{}.
\newblock \showarticletitle{Counterfactual Augmentation for Robust Authorship Representation Learning}. In \bibinfo{booktitle}{\emph{Proceedings of the 47th International ACM SIGIR Conference on Research and Development in Information Retrieval}} (Washington DC, USA) \emph{(\bibinfo{series}{SIGIR '24})}. \bibinfo{publisher}{Association for Computing Machinery}, \bibinfo{address}{New York, NY, USA}, \bibinfo{pages}{2347–2351}.
\newblock
\showISBNx{9798400704314}
\href{https://doi.org/10.1145/3626772.3657956}{doi:\nolinkurl{10.1145/3626772.3657956}}


\bibitem[Merity et~al\mbox{.}(2017)]%
        {merity2017pointer}
\bibfield{author}{\bibinfo{person}{Stephen Merity}, \bibinfo{person}{Caiming Xiong}, \bibinfo{person}{James Bradbury}, {and} \bibinfo{person}{Richard Socher}.} \bibinfo{year}{2017}\natexlab{}.
\newblock \showarticletitle{Pointer Sentinel Mixture Models}. In \bibinfo{booktitle}{\emph{International Conference on Learning Representations}}.
\newblock
\urldef\tempurl%
\url{https://openreview.net/forum?id=Byj72udxe}
\showURL{%
\tempurl}


\bibitem[Mikolov et~al\mbox{.}(2013)]%
        {mikolov2013distributedrepresentationswordsphrases}
\bibfield{author}{\bibinfo{person}{Tomas Mikolov}, \bibinfo{person}{Ilya Sutskever}, \bibinfo{person}{Kai Chen}, \bibinfo{person}{Greg Corrado}, {and} \bibinfo{person}{Jeffrey Dean}.} \bibinfo{year}{2013}\natexlab{}.
\newblock \bibinfo{title}{Distributed Representations of Words and Phrases and their Compositionality}.
\newblock
\showeprint[arxiv]{1310.4546}~[cs.CL]
\urldef\tempurl%
\url{https://arxiv.org/abs/1310.4546}
\showURL{%
\tempurl}


\bibitem[Muennighoff et~al\mbox{.}(2024)]%
        {muennighoff2024generativerepresentationalinstructiontuning}
\bibfield{author}{\bibinfo{person}{Niklas Muennighoff}, \bibinfo{person}{Hongjin Su}, \bibinfo{person}{Liang Wang}, \bibinfo{person}{Nan Yang}, \bibinfo{person}{Furu Wei}, \bibinfo{person}{Tao Yu}, \bibinfo{person}{Amanpreet Singh}, {and} \bibinfo{person}{Douwe Kiela}.} \bibinfo{year}{2024}\natexlab{}.
\newblock \bibinfo{title}{Generative Representational Instruction Tuning}.
\newblock
\showeprint[arxiv]{2402.09906}~[cs.CL]
\urldef\tempurl%
\url{https://arxiv.org/abs/2402.09906}
\showURL{%
\tempurl}


\bibitem[Muennighoff et~al\mbox{.}(2023)]%
        {muennighoff-etal-2023-mteb}
\bibfield{author}{\bibinfo{person}{Niklas Muennighoff}, \bibinfo{person}{Nouamane Tazi}, \bibinfo{person}{Loic Magne}, {and} \bibinfo{person}{Nils Reimers}.} \bibinfo{year}{2023}\natexlab{}.
\newblock \showarticletitle{{MTEB}: Massive Text Embedding Benchmark}. In \bibinfo{booktitle}{\emph{Proceedings of the 17th Conference of the European Chapter of the Association for Computational Linguistics}}, \bibfield{editor}{\bibinfo{person}{Andreas Vlachos} {and} \bibinfo{person}{Isabelle Augenstein}} (Eds.). \bibinfo{publisher}{Association for Computational Linguistics}, \bibinfo{address}{Dubrovnik, Croatia}, \bibinfo{pages}{2014--2037}.
\newblock
\href{https://doi.org/10.18653/v1/2023.eacl-main.148}{doi:\nolinkurl{10.18653/v1/2023.eacl-main.148}}


\bibitem[Ni et~al\mbox{.}(2021a)]%
        {ni2021large}
\bibfield{author}{\bibinfo{person}{Jianmo Ni}, \bibinfo{person}{Chen Qu}, \bibinfo{person}{Jing Lu}, \bibinfo{person}{Zhuyun Dai}, \bibinfo{person}{Gustavo~Hern{\'a}ndez {\'A}brego}, \bibinfo{person}{Ji Ma}, \bibinfo{person}{Vincent~Y Zhao}, \bibinfo{person}{Yi Luan}, \bibinfo{person}{Keith~B Hall}, \bibinfo{person}{Ming-Wei Chang}, {et~al\mbox{.}}} \bibinfo{year}{2021}\natexlab{a}.
\newblock \showarticletitle{Large dual encoders are generalizable retrievers}.
\newblock \bibinfo{journal}{\emph{arXiv preprint arXiv:2112.07899}} (\bibinfo{year}{2021}).
\newblock


\bibitem[Ni et~al\mbox{.}(2021b)]%
        {ni2021largedualencodersgeneralizable}
\bibfield{author}{\bibinfo{person}{Jianmo Ni}, \bibinfo{person}{Chen Qu}, \bibinfo{person}{Jing Lu}, \bibinfo{person}{Zhuyun Dai}, \bibinfo{person}{Gustavo~Hernández Ábrego}, \bibinfo{person}{Ji Ma}, \bibinfo{person}{Vincent~Y. Zhao}, \bibinfo{person}{Yi Luan}, \bibinfo{person}{Keith~B. Hall}, \bibinfo{person}{Ming-Wei Chang}, {and} \bibinfo{person}{Yinfei Yang}.} \bibinfo{year}{2021}\natexlab{b}.
\newblock \bibinfo{title}{Large Dual Encoders Are Generalizable Retrievers}.
\newblock
\showeprint[arxiv]{2112.07899}~[cs.IR]
\urldef\tempurl%
\url{https://arxiv.org/abs/2112.07899}
\showURL{%
\tempurl}


\bibitem[Ni et~al\mbox{.}(2021c)]%
        {ni2021sentencet5scalablesentenceencoders}
\bibfield{author}{\bibinfo{person}{Jianmo Ni}, \bibinfo{person}{Gustavo~Hernández Ábrego}, \bibinfo{person}{Noah Constant}, \bibinfo{person}{Ji Ma}, \bibinfo{person}{Keith~B. Hall}, \bibinfo{person}{Daniel Cer}, {and} \bibinfo{person}{Yinfei Yang}.} \bibinfo{year}{2021}\natexlab{c}.
\newblock \bibinfo{title}{Sentence-T5: Scalable Sentence Encoders from Pre-trained Text-to-Text Models}.
\newblock
\showeprint[arxiv]{2108.08877}~[cs.CL]
\urldef\tempurl%
\url{https://arxiv.org/abs/2108.08877}
\showURL{%
\tempurl}


\bibitem[Pires et~al\mbox{.}(2019)]%
        {pires-etal-2019-multilingual}
\bibfield{author}{\bibinfo{person}{Telmo Pires}, \bibinfo{person}{Eva Schlinger}, {and} \bibinfo{person}{Dan Garrette}.} \bibinfo{year}{2019}\natexlab{}.
\newblock \showarticletitle{How Multilingual is Multilingual {BERT}?}. In \bibinfo{booktitle}{\emph{Proceedings of the 57th Annual Meeting of the Association for Computational Linguistics}}, \bibfield{editor}{\bibinfo{person}{Anna Korhonen}, \bibinfo{person}{David Traum}, {and} \bibinfo{person}{Llu{\'\i}s M{\`a}rquez}} (Eds.). \bibinfo{publisher}{Association for Computational Linguistics}, \bibinfo{address}{Florence, Italy}, \bibinfo{pages}{4996--5001}.
\newblock
\href{https://doi.org/10.18653/v1/P19-1493}{doi:\nolinkurl{10.18653/v1/P19-1493}}


\bibitem[Rajpurkar et~al\mbox{.}(2016)]%
        {rajpurkar-etal-2016-squad}
\bibfield{author}{\bibinfo{person}{Pranav Rajpurkar}, \bibinfo{person}{Jian Zhang}, \bibinfo{person}{Konstantin Lopyrev}, {and} \bibinfo{person}{Percy Liang}.} \bibinfo{year}{2016}\natexlab{}.
\newblock \showarticletitle{{SQ}u{AD}: 100,000+ Questions for Machine Comprehension of Text}. In \bibinfo{booktitle}{\emph{Proceedings of the 2016 Conference on Empirical Methods in Natural Language Processing}}, \bibfield{editor}{\bibinfo{person}{Jian Su}, \bibinfo{person}{Kevin Duh}, {and} \bibinfo{person}{Xavier Carreras}} (Eds.). \bibinfo{publisher}{Association for Computational Linguistics}, \bibinfo{address}{Austin, Texas}, \bibinfo{pages}{2383--2392}.
\newblock
\href{https://doi.org/10.18653/v1/D16-1264}{doi:\nolinkurl{10.18653/v1/D16-1264}}


\bibitem[Ramesh et~al\mbox{.}(2022)]%
        {DBLP:journals/tacl/RameshDBJASSDJK22}
\bibfield{author}{\bibinfo{person}{Gowtham Ramesh}, \bibinfo{person}{Sumanth Doddapaneni}, \bibinfo{person}{Aravinth Bheemaraj}, \bibinfo{person}{Mayank Jobanputra}, \bibinfo{person}{Raghavan AK}, \bibinfo{person}{Ajitesh Sharma}, \bibinfo{person}{Sujit Sahoo}, \bibinfo{person}{Harshita Diddee}, \bibinfo{person}{Mahalakshmi J}, \bibinfo{person}{Divyanshu Kakwani}, \bibinfo{person}{Navneet Kumar}, \bibinfo{person}{Aswin Pradeep}, \bibinfo{person}{Srihari Nagaraj}, \bibinfo{person}{Deepak Kumar}, \bibinfo{person}{Vivek Raghavan}, \bibinfo{person}{Anoop Kunchukuttan}, \bibinfo{person}{Pratyush Kumar}, {and} \bibinfo{person}{Mitesh~Shantadevi Khapra}.} \bibinfo{year}{2022}\natexlab{}.
\newblock \showarticletitle{Samanantar: The Largest Publicly Available Parallel Corpora Collection for 11 Indic Languages}.
\newblock \bibinfo{journal}{\emph{Trans. Assoc. Comput. Linguistics}}  \bibinfo{volume}{10} (\bibinfo{year}{2022}), \bibinfo{pages}{145--162}.
\newblock
\href{https://doi.org/10.1162/TACL\_A\_00452}{doi:\nolinkurl{10.1162/TACL\_A\_00452}}


\bibitem[Reimers et~al\mbox{.}(2016)]%
        {reimers-etal-2016-task}
\bibfield{author}{\bibinfo{person}{Nils Reimers}, \bibinfo{person}{Philip Beyer}, {and} \bibinfo{person}{Iryna Gurevych}.} \bibinfo{year}{2016}\natexlab{}.
\newblock \showarticletitle{Task-Oriented Intrinsic Evaluation of Semantic Textual Similarity}. In \bibinfo{booktitle}{\emph{Proceedings of {COLING} 2016, the 26th International Conference on Computational Linguistics: Technical Papers}}, \bibfield{editor}{\bibinfo{person}{Yuji Matsumoto} {and} \bibinfo{person}{Rashmi Prasad}} (Eds.). \bibinfo{publisher}{The COLING 2016 Organizing Committee}, \bibinfo{address}{Osaka, Japan}, \bibinfo{pages}{87--96}.
\newblock
\urldef\tempurl%
\url{https://aclanthology.org/C16-1009}
\showURL{%
\tempurl}


\bibitem[Reimers and Gurevych(2019)]%
        {reimers-gurevych-2019-sentence}
\bibfield{author}{\bibinfo{person}{Nils Reimers} {and} \bibinfo{person}{Iryna Gurevych}.} \bibinfo{year}{2019}\natexlab{}.
\newblock \showarticletitle{Sentence-{BERT}: Sentence Embeddings using {S}iamese {BERT}-Networks}. In \bibinfo{booktitle}{\emph{Proceedings of the 2019 Conference on Empirical Methods in Natural Language Processing and the 9th International Joint Conference on Natural Language Processing (EMNLP-IJCNLP)}}, \bibfield{editor}{\bibinfo{person}{Kentaro Inui}, \bibinfo{person}{Jing Jiang}, \bibinfo{person}{Vincent Ng}, {and} \bibinfo{person}{Xiaojun Wan}} (Eds.). \bibinfo{publisher}{Association for Computational Linguistics}, \bibinfo{address}{Hong Kong, China}, \bibinfo{pages}{3982--3992}.
\newblock
\href{https://doi.org/10.18653/v1/D19-1410}{doi:\nolinkurl{10.18653/v1/D19-1410}}


\bibitem[Rivera-Soto et~al\mbox{.}(2021)]%
        {rivera-soto-etal-2021-learning}
\bibfield{author}{\bibinfo{person}{Rafael~A. Rivera-Soto}, \bibinfo{person}{Olivia~Elizabeth Miano}, \bibinfo{person}{Juanita Ordonez}, \bibinfo{person}{Barry~Y. Chen}, \bibinfo{person}{Aleem Khan}, \bibinfo{person}{Marcus Bishop}, {and} \bibinfo{person}{Nicholas Andrews}.} \bibinfo{year}{2021}\natexlab{}.
\newblock \showarticletitle{Learning Universal Authorship Representations}. In \bibinfo{booktitle}{\emph{Proceedings of the 2021 Conference on Empirical Methods in Natural Language Processing}}, \bibfield{editor}{\bibinfo{person}{Marie-Francine Moens}, \bibinfo{person}{Xuanjing Huang}, \bibinfo{person}{Lucia Specia}, {and} \bibinfo{person}{Scott Wen-tau Yih}} (Eds.). \bibinfo{publisher}{Association for Computational Linguistics}, \bibinfo{address}{Online and Punta Cana, Dominican Republic}, \bibinfo{pages}{913--919}.
\newblock
\href{https://doi.org/10.18653/v1/2021.emnlp-main.70}{doi:\nolinkurl{10.18653/v1/2021.emnlp-main.70}}


\bibitem[Robertson et~al\mbox{.}(2009)]%
        {robertson2009probabilistic}
\bibfield{author}{\bibinfo{person}{Stephen Robertson}, \bibinfo{person}{Hugo Zaragoza}, {et~al\mbox{.}}} \bibinfo{year}{2009}\natexlab{}.
\newblock \showarticletitle{The probabilistic relevance framework: BM25 and beyond}.
\newblock \bibinfo{journal}{\emph{Foundations and Trends{\textregistered} in Information Retrieval}} \bibinfo{volume}{3}, \bibinfo{number}{4} (\bibinfo{year}{2009}), \bibinfo{pages}{333--389}.
\newblock


\bibitem[Rosenberg and Hirschberg(2007)]%
        {rosenberg-hirschberg-2007-v}
\bibfield{author}{\bibinfo{person}{Andrew Rosenberg} {and} \bibinfo{person}{Julia Hirschberg}.} \bibinfo{year}{2007}\natexlab{}.
\newblock \showarticletitle{{V}-Measure: A Conditional Entropy-Based External Cluster Evaluation Measure}. In \bibinfo{booktitle}{\emph{Proceedings of the 2007 Joint Conference on Empirical Methods in Natural Language Processing and Computational Natural Language Learning ({EMNLP}-{C}o{NLL})}}, \bibfield{editor}{\bibinfo{person}{Jason Eisner}} (Ed.). \bibinfo{publisher}{Association for Computational Linguistics}, \bibinfo{address}{Prague, Czech Republic}, \bibinfo{pages}{410--420}.
\newblock
\urldef\tempurl%
\url{https://aclanthology.org/D07-1043}
\showURL{%
\tempurl}


\bibitem[Shi et~al\mbox{.}(2021)]%
        {shi-etal-2021-cross}
\bibfield{author}{\bibinfo{person}{Peng Shi}, \bibinfo{person}{Rui Zhang}, \bibinfo{person}{He Bai}, {and} \bibinfo{person}{Jimmy Lin}.} \bibinfo{year}{2021}\natexlab{}.
\newblock \showarticletitle{Cross-Lingual Training of Dense Retrievers for Document Retrieval}. In \bibinfo{booktitle}{\emph{Proceedings of the 1st Workshop on Multilingual Representation Learning}}, \bibfield{editor}{\bibinfo{person}{Duygu Ataman}, \bibinfo{person}{Alexandra Birch}, \bibinfo{person}{Alexis Conneau}, \bibinfo{person}{Orhan Firat}, \bibinfo{person}{Sebastian Ruder}, {and} \bibinfo{person}{Gozde~Gul Sahin}} (Eds.). \bibinfo{publisher}{Association for Computational Linguistics}, \bibinfo{address}{Punta Cana, Dominican Republic}, \bibinfo{pages}{251--253}.
\newblock
\href{https://doi.org/10.18653/v1/2021.mrl-1.24}{doi:\nolinkurl{10.18653/v1/2021.mrl-1.24}}


\bibitem[Sturua et~al\mbox{.}(2024)]%
        {sturua2024jinaembeddingsv3multilingualembeddingstask}
\bibfield{author}{\bibinfo{person}{Saba Sturua}, \bibinfo{person}{Isabelle Mohr}, \bibinfo{person}{Mohammad~Kalim Akram}, \bibinfo{person}{Michael Günther}, \bibinfo{person}{Bo Wang}, \bibinfo{person}{Markus Krimmel}, \bibinfo{person}{Feng Wang}, \bibinfo{person}{Georgios Mastrapas}, \bibinfo{person}{Andreas Koukounas}, \bibinfo{person}{Nan Wang}, {and} \bibinfo{person}{Han Xiao}.} \bibinfo{year}{2024}\natexlab{}.
\newblock \bibinfo{title}{jina-embeddings-v3: Multilingual Embeddings With Task LoRA}.
\newblock
\showeprint[arxiv]{2409.10173}~[cs.CL]
\urldef\tempurl%
\url{https://arxiv.org/abs/2409.10173}
\showURL{%
\tempurl}


\bibitem[Thakur et~al\mbox{.}(2024)]%
        {thakur2024leveragingllmssynthesizingtraining}
\bibfield{author}{\bibinfo{person}{Nandan Thakur}, \bibinfo{person}{Jianmo Ni}, \bibinfo{person}{Gustavo~Hernández Ábrego}, \bibinfo{person}{John Wieting}, \bibinfo{person}{Jimmy Lin}, {and} \bibinfo{person}{Daniel Cer}.} \bibinfo{year}{2024}\natexlab{}.
\newblock \bibinfo{title}{Leveraging LLMs for Synthesizing Training Data Across Many Languages in Multilingual Dense Retrieval}.
\newblock
\showeprint[arxiv]{2311.05800}~[cs.IR]
\urldef\tempurl%
\url{https://arxiv.org/abs/2311.05800}
\showURL{%
\tempurl}


\bibitem[Thorne et~al\mbox{.}(2018)]%
        {thorne-etal-2018-fever}
\bibfield{author}{\bibinfo{person}{James Thorne}, \bibinfo{person}{Andreas Vlachos}, \bibinfo{person}{Christos Christodoulopoulos}, {and} \bibinfo{person}{Arpit Mittal}.} \bibinfo{year}{2018}\natexlab{}.
\newblock \showarticletitle{{FEVER}: a Large-scale Dataset for Fact Extraction and {VER}ification}. In \bibinfo{booktitle}{\emph{Proceedings of the 2018 Conference of the North {A}merican Chapter of the Association for Computational Linguistics: Human Language Technologies, Volume 1 (Long Papers)}}, \bibfield{editor}{\bibinfo{person}{Marilyn Walker}, \bibinfo{person}{Heng Ji}, {and} \bibinfo{person}{Amanda Stent}} (Eds.). \bibinfo{publisher}{Association for Computational Linguistics}, \bibinfo{address}{New Orleans, Louisiana}, \bibinfo{pages}{809--819}.
\newblock
\href{https://doi.org/10.18653/v1/N18-1074}{doi:\nolinkurl{10.18653/v1/N18-1074}}


\bibitem[Touvron et~al\mbox{.}(2023)]%
        {touvron2023llamaopenefficientfoundation}
\bibfield{author}{\bibinfo{person}{Hugo Touvron}, \bibinfo{person}{Thibaut Lavril}, \bibinfo{person}{Gautier Izacard}, \bibinfo{person}{Xavier Martinet}, \bibinfo{person}{Marie-Anne Lachaux}, \bibinfo{person}{Timothée Lacroix}, \bibinfo{person}{Baptiste Rozière}, \bibinfo{person}{Naman Goyal}, \bibinfo{person}{Eric Hambro}, \bibinfo{person}{Faisal Azhar}, \bibinfo{person}{Aurelien Rodriguez}, \bibinfo{person}{Armand Joulin}, \bibinfo{person}{Edouard Grave}, {and} \bibinfo{person}{Guillaume Lample}.} \bibinfo{year}{2023}\natexlab{}.
\newblock \bibinfo{title}{LLaMA: Open and Efficient Foundation Language Models}.
\newblock
\showeprint[arxiv]{2302.13971}~[cs.CL]
\urldef\tempurl%
\url{https://arxiv.org/abs/2302.13971}
\showURL{%
\tempurl}


\bibitem[Wachsmuth et~al\mbox{.}(2018)]%
        {wachsmuth-etal-2018-retrieval}
\bibfield{author}{\bibinfo{person}{Henning Wachsmuth}, \bibinfo{person}{Shahbaz Syed}, {and} \bibinfo{person}{Benno Stein}.} \bibinfo{year}{2018}\natexlab{}.
\newblock \showarticletitle{Retrieval of the Best Counterargument without Prior Topic Knowledge}. In \bibinfo{booktitle}{\emph{Proceedings of the 56th Annual Meeting of the Association for Computational Linguistics (Volume 1: Long Papers)}}, \bibfield{editor}{\bibinfo{person}{Iryna Gurevych} {and} \bibinfo{person}{Yusuke Miyao}} (Eds.). \bibinfo{publisher}{Association for Computational Linguistics}, \bibinfo{address}{Melbourne, Australia}, \bibinfo{pages}{241--251}.
\newblock
\href{https://doi.org/10.18653/v1/P18-1023}{doi:\nolinkurl{10.18653/v1/P18-1023}}


\bibitem[Wang et~al\mbox{.}(2023)]%
        {wang2023instructretro}
\bibfield{author}{\bibinfo{person}{Boxin Wang}, \bibinfo{person}{Wei Ping}, \bibinfo{person}{Lawrence McAfee}, \bibinfo{person}{Peng Xu}, \bibinfo{person}{Bo Li}, \bibinfo{person}{Mohammad Shoeybi}, {and} \bibinfo{person}{Bryan Catanzaro}.} \bibinfo{year}{2023}\natexlab{}.
\newblock \showarticletitle{Instructretro: Instruction tuning post retrieval-augmented pretraining}.
\newblock \bibinfo{journal}{\emph{arXiv preprint arXiv:2310.07713}} (\bibinfo{year}{2023}).
\newblock


\bibitem[Wang et~al\mbox{.}(2024a)]%
        {wang2024textembeddingsweaklysupervisedcontrastive}
\bibfield{author}{\bibinfo{person}{Liang Wang}, \bibinfo{person}{Nan Yang}, \bibinfo{person}{Xiaolong Huang}, \bibinfo{person}{Binxing Jiao}, \bibinfo{person}{Linjun Yang}, \bibinfo{person}{Daxin Jiang}, \bibinfo{person}{Rangan Majumder}, {and} \bibinfo{person}{Furu Wei}.} \bibinfo{year}{2024}\natexlab{a}.
\newblock \bibinfo{title}{Text Embeddings by Weakly-Supervised Contrastive Pre-training}.
\newblock
\showeprint[arxiv]{2212.03533}~[cs.CL]
\urldef\tempurl%
\url{https://arxiv.org/abs/2212.03533}
\showURL{%
\tempurl}


\bibitem[Wang et~al\mbox{.}(2024b)]%
        {wang-etal-2024-improving-text}
\bibfield{author}{\bibinfo{person}{Liang Wang}, \bibinfo{person}{Nan Yang}, \bibinfo{person}{Xiaolong Huang}, \bibinfo{person}{Linjun Yang}, \bibinfo{person}{Rangan Majumder}, {and} \bibinfo{person}{Furu Wei}.} \bibinfo{year}{2024}\natexlab{b}.
\newblock \showarticletitle{Improving Text Embeddings with Large Language Models}. In \bibinfo{booktitle}{\emph{Proceedings of the 62nd Annual Meeting of the Association for Computational Linguistics (Volume 1: Long Papers)}}, \bibfield{editor}{\bibinfo{person}{Lun-Wei Ku}, \bibinfo{person}{Andre Martins}, {and} \bibinfo{person}{Vivek Srikumar}} (Eds.). \bibinfo{publisher}{Association for Computational Linguistics}, \bibinfo{address}{Bangkok, Thailand}, \bibinfo{pages}{11897--11916}.
\newblock
\href{https://doi.org/10.18653/v1/2024.acl-long.642}{doi:\nolinkurl{10.18653/v1/2024.acl-long.642}}


\bibitem[Wang et~al\mbox{.}(2024c)]%
        {wang2024multilinguale5textembeddings}
\bibfield{author}{\bibinfo{person}{Liang Wang}, \bibinfo{person}{Nan Yang}, \bibinfo{person}{Xiaolong Huang}, \bibinfo{person}{Linjun Yang}, \bibinfo{person}{Rangan Majumder}, {and} \bibinfo{person}{Furu Wei}.} \bibinfo{year}{2024}\natexlab{c}.
\newblock \bibinfo{title}{Multilingual E5 Text Embeddings: A Technical Report}.
\newblock
\showeprint[arxiv]{2402.05672}~[cs.CL]
\urldef\tempurl%
\url{https://arxiv.org/abs/2402.05672}
\showURL{%
\tempurl}


\bibitem[Wang et~al\mbox{.}(2024d)]%
        {wang2024multilingual}
\bibfield{author}{\bibinfo{person}{Liang Wang}, \bibinfo{person}{Nan Yang}, \bibinfo{person}{Xiaolong Huang}, \bibinfo{person}{Linjun Yang}, \bibinfo{person}{Rangan Majumder}, {and} \bibinfo{person}{Furu Wei}.} \bibinfo{year}{2024}\natexlab{d}.
\newblock \showarticletitle{Multilingual E5 Text Embeddings: A Technical Report}.
\newblock \bibinfo{journal}{\emph{arXiv preprint arXiv:2402.05672}} (\bibinfo{year}{2024}).
\newblock


\bibitem[Wang et~al\mbox{.}(2020)]%
        {wang-etal-2020-extending}
\bibfield{author}{\bibinfo{person}{Zihan Wang}, \bibinfo{person}{Karthikeyan K}, \bibinfo{person}{Stephen Mayhew}, {and} \bibinfo{person}{Dan Roth}.} \bibinfo{year}{2020}\natexlab{}.
\newblock \showarticletitle{Extending Multilingual {BERT} to Low-Resource Languages}. In \bibinfo{booktitle}{\emph{Findings of the Association for Computational Linguistics: EMNLP 2020}}, \bibfield{editor}{\bibinfo{person}{Trevor Cohn}, \bibinfo{person}{Yulan He}, {and} \bibinfo{person}{Yang Liu}} (Eds.). \bibinfo{publisher}{Association for Computational Linguistics}, \bibinfo{address}{Online}, \bibinfo{pages}{2649--2656}.
\newblock
\href{https://doi.org/10.18653/v1/2020.findings-emnlp.240}{doi:\nolinkurl{10.18653/v1/2020.findings-emnlp.240}}


\bibitem[Winata et~al\mbox{.}(2024)]%
        {winata2024minersmultilinguallanguagemodels}
\bibfield{author}{\bibinfo{person}{Genta~Indra Winata}, \bibinfo{person}{Ruochen Zhang}, {and} \bibinfo{person}{David~Ifeoluwa Adelani}.} \bibinfo{year}{2024}\natexlab{}.
\newblock \bibinfo{title}{MINERS: Multilingual Language Models as Semantic Retrievers}.
\newblock
\showeprint[arxiv]{2406.07424}~[cs.CL]
\urldef\tempurl%
\url{https://arxiv.org/abs/2406.07424}
\showURL{%
\tempurl}


\bibitem[Wolf et~al\mbox{.}(2020)]%
        {wolf-etal-2020-transformers}
\bibfield{author}{\bibinfo{person}{Thomas Wolf}, \bibinfo{person}{Lysandre Debut}, \bibinfo{person}{Victor Sanh}, \bibinfo{person}{Julien Chaumond}, \bibinfo{person}{Clement Delangue}, \bibinfo{person}{Anthony Moi}, \bibinfo{person}{Pierric Cistac}, \bibinfo{person}{Tim Rault}, \bibinfo{person}{Remi Louf}, \bibinfo{person}{Morgan Funtowicz}, \bibinfo{person}{Joe Davison}, \bibinfo{person}{Sam Shleifer}, \bibinfo{person}{Patrick von Platen}, \bibinfo{person}{Clara Ma}, \bibinfo{person}{Yacine Jernite}, \bibinfo{person}{Julien Plu}, \bibinfo{person}{Canwen Xu}, \bibinfo{person}{Teven Le~Scao}, \bibinfo{person}{Sylvain Gugger}, \bibinfo{person}{Mariama Drame}, \bibinfo{person}{Quentin Lhoest}, {and} \bibinfo{person}{Alexander Rush}.} \bibinfo{year}{2020}\natexlab{}.
\newblock \showarticletitle{Transformers: State-of-the-Art Natural Language Processing}. In \bibinfo{booktitle}{\emph{Proceedings of the 2020 Conference on Empirical Methods in Natural Language Processing: System Demonstrations}}, \bibfield{editor}{\bibinfo{person}{Qun Liu} {and} \bibinfo{person}{David Schlangen}} (Eds.). \bibinfo{publisher}{Association for Computational Linguistics}, \bibinfo{address}{Online}, \bibinfo{pages}{38--45}.
\newblock
\href{https://doi.org/10.18653/v1/2020.emnlp-demos.6}{doi:\nolinkurl{10.18653/v1/2020.emnlp-demos.6}}


\bibitem[Xiao et~al\mbox{.}(2023)]%
        {bge_embedding}
\bibfield{author}{\bibinfo{person}{Shitao Xiao}, \bibinfo{person}{Zheng Liu}, \bibinfo{person}{Peitian Zhang}, {and} \bibinfo{person}{Niklas Muennighoff}.} \bibinfo{year}{2023}\natexlab{}.
\newblock \bibinfo{title}{C-Pack: Packaged Resources To Advance General Chinese Embedding}.
\newblock
\showeprint[arxiv]{2309.07597}~[cs.CL]


\bibitem[Yang et~al\mbox{.}(2018)]%
        {yang-etal-2018-hotpotqa}
\bibfield{author}{\bibinfo{person}{Zhilin Yang}, \bibinfo{person}{Peng Qi}, \bibinfo{person}{Saizheng Zhang}, \bibinfo{person}{Yoshua Bengio}, \bibinfo{person}{William Cohen}, \bibinfo{person}{Ruslan Salakhutdinov}, {and} \bibinfo{person}{Christopher~D. Manning}.} \bibinfo{year}{2018}\natexlab{}.
\newblock \showarticletitle{{H}otpot{QA}: A Dataset for Diverse, Explainable Multi-hop Question Answering}. In \bibinfo{booktitle}{\emph{Proceedings of the 2018 Conference on Empirical Methods in Natural Language Processing}}, \bibfield{editor}{\bibinfo{person}{Ellen Riloff}, \bibinfo{person}{David Chiang}, \bibinfo{person}{Julia Hockenmaier}, {and} \bibinfo{person}{Jun{'}ichi Tsujii}} (Eds.). \bibinfo{publisher}{Association for Computational Linguistics}, \bibinfo{address}{Brussels, Belgium}, \bibinfo{pages}{2369--2380}.
\newblock
\href{https://doi.org/10.18653/v1/D18-1259}{doi:\nolinkurl{10.18653/v1/D18-1259}}


\bibitem[Yoon et~al\mbox{.}(2024)]%
        {yoon-etal-2024-langbridge}
\bibfield{author}{\bibinfo{person}{Dongkeun Yoon}, \bibinfo{person}{Joel Jang}, \bibinfo{person}{Sungdong Kim}, \bibinfo{person}{Seungone Kim}, \bibinfo{person}{Sheikh Shafayat}, {and} \bibinfo{person}{Minjoon Seo}.} \bibinfo{year}{2024}\natexlab{}.
\newblock \showarticletitle{{L}ang{B}ridge: Multilingual Reasoning Without Multilingual Supervision}. In \bibinfo{booktitle}{\emph{Proceedings of the 62nd Annual Meeting of the Association for Computational Linguistics (Volume 1: Long Papers)}}, \bibfield{editor}{\bibinfo{person}{Lun-Wei Ku}, \bibinfo{person}{Andre Martins}, {and} \bibinfo{person}{Vivek Srikumar}} (Eds.). \bibinfo{publisher}{Association for Computational Linguistics}, \bibinfo{address}{Bangkok, Thailand}, \bibinfo{pages}{7502--7522}.
\newblock
\href{https://doi.org/10.18653/v1/2024.acl-long.405}{doi:\nolinkurl{10.18653/v1/2024.acl-long.405}}


\bibitem[Zhang and Misra(2022)]%
        {zhang-misra-2022-machine}
\bibfield{author}{\bibinfo{person}{Bryan Zhang} {and} \bibinfo{person}{Amita Misra}.} \bibinfo{year}{2022}\natexlab{}.
\newblock \showarticletitle{Machine translation impact in {E}-commerce multilingual search}. In \bibinfo{booktitle}{\emph{Proceedings of the 2022 Conference on Empirical Methods in Natural Language Processing: Industry Track}}, \bibfield{editor}{\bibinfo{person}{Yunyao Li} {and} \bibinfo{person}{Angeliki Lazaridou}} (Eds.). \bibinfo{publisher}{Association for Computational Linguistics}, \bibinfo{address}{Abu Dhabi, UAE}, \bibinfo{pages}{99--109}.
\newblock
\href{https://doi.org/10.18653/v1/2022.emnlp-industry.8}{doi:\nolinkurl{10.18653/v1/2022.emnlp-industry.8}}


\bibitem[Zhang et~al\mbox{.}(2023)]%
        {zhang2023language}
\bibfield{author}{\bibinfo{person}{Xin Zhang}, \bibinfo{person}{Zehan Li}, \bibinfo{person}{Yanzhao Zhang}, \bibinfo{person}{Dingkun Long}, \bibinfo{person}{Pengjun Xie}, \bibinfo{person}{Meishan Zhang}, {and} \bibinfo{person}{Min Zhang}.} \bibinfo{year}{2023}\natexlab{}.
\newblock \showarticletitle{Language Models are Universal Embedders}.
\newblock \bibinfo{journal}{\emph{arXiv preprint arXiv:2310.08232}} (\bibinfo{year}{2023}).
\newblock


\bibitem[Zhang et~al\mbox{.}(2024)]%
        {zhang2024mgte}
\bibfield{author}{\bibinfo{person}{Xin Zhang}, \bibinfo{person}{Yanzhao Zhang}, \bibinfo{person}{Dingkun Long}, \bibinfo{person}{Wen Xie}, \bibinfo{person}{Ziqi Dai}, \bibinfo{person}{Jialong Tang}, \bibinfo{person}{Huan Lin}, \bibinfo{person}{Baosong Yang}, \bibinfo{person}{Pengjun Xie}, \bibinfo{person}{Fei Huang}, \bibinfo{person}{Meishan Zhang}, \bibinfo{person}{Wenjie Li}, {and} \bibinfo{person}{Min Zhang}.} \bibinfo{year}{2024}\natexlab{}.
\newblock \bibinfo{title}{mGTE: Generalized Long-Context Text Representation and Reranking Models for Multilingual Text Retrieval}.
\newblock
\showeprint[arxiv]{2407.19669}~[cs.CL]
\urldef\tempurl%
\url{https://arxiv.org/abs/2407.19669}
\showURL{%
\tempurl}


\bibitem[Zhao et~al\mbox{.}(2023)]%
        {zhao2023pytorchfsdpexperiencesscaling}
\bibfield{author}{\bibinfo{person}{Yanli Zhao}, \bibinfo{person}{Andrew Gu}, \bibinfo{person}{Rohan Varma}, \bibinfo{person}{Liang Luo}, \bibinfo{person}{Chien-Chin Huang}, \bibinfo{person}{Min Xu}, \bibinfo{person}{Less Wright}, \bibinfo{person}{Hamid Shojanazeri}, \bibinfo{person}{Myle Ott}, \bibinfo{person}{Sam Shleifer}, \bibinfo{person}{Alban Desmaison}, \bibinfo{person}{Can Balioglu}, \bibinfo{person}{Pritam Damania}, \bibinfo{person}{Bernard Nguyen}, \bibinfo{person}{Geeta Chauhan}, \bibinfo{person}{Yuchen Hao}, \bibinfo{person}{Ajit Mathews}, {and} \bibinfo{person}{Shen Li}.} \bibinfo{year}{2023}\natexlab{}.
\newblock \bibinfo{title}{PyTorch FSDP: Experiences on Scaling Fully Sharded Data Parallel}.
\newblock
\showeprint[arxiv]{2304.11277}~[cs.DC]
\urldef\tempurl%
\url{https://arxiv.org/abs/2304.11277}
\showURL{%
\tempurl}


\end{thebibliography}

\appendix
\section{Detailed Results}

In this section, we provide detailed results of LUSIFER and E5-Mistral on all benchmark datasets for each language. 

\begin{table}[!htp]
  \centering
  \small
  \resizebox{\columnwidth}{!}{
  \begin{tabular}{l|c|c}
     \toprule
      Es Datasets & E5-Mistral & LUSIFER \\ 
      \midrule
      AmazonReviewsClassification & 42.69 & 50.41 \\ 
      MassiveIntentClassification & 69.67 & 68.93 \\ 
      MassiveScenarioClassification & 74.63 & 73.41 \\ 
      MTOPIntentClassification & 72.16 & 80.13 \\ 
      MultilingualSentimentClassification & 87.91 & 91.01 \\ 
      TweetSentimentClassification & 49.73 & 58.55 \\ 
      SpanishNewsClassification & 89.5 & 87.81 \\ 
      PawsXPairClassification & 61.19 & 62.82 \\ 
      XNLI & 77.34 & 60.49 \\ 
      SpanishNewsClusteringP2P & 42.28 & 43.85 \\ 
      MLSUMClusteringP2P & 47.54 & 44.36 \\ 
      MLSUMClusteringS2S & 47.11 & 41.56 \\ 
      SIB200ClusteringS2S & 31.01 & 44.42 \\ 
      MultiEURLEXMultilabelClassification & 6.16 & 3.87 \\ 
      BelebeleRetrieval & 83.92 & 81.4 \\ 
      MintakaRetrieval & 48.77 & 18.17 \\ 
      STS17 & 87.18 & 80.84 \\ 
      STS22 & 71.79 & 70.66 \\ 
      STSBenchmarkMultilingualSTS & 84.31 & 79.89 \\ 
      \midrule
      Avg. & 61.84 & 60.14 \\ 
      \bottomrule
  \end{tabular}
  }
  \caption{Detailed results of E5-Mistral and LUSIFER on the Spanish benchmark datasets.}
\end{table}

\begin{table}[!htp]
  \centering
  \small
  \resizebox{\columnwidth}{!}{
  \begin{tabular}{l|c|c}
    \toprule
    \text{En Datasets} & E5-Mistral & LUSIFER \\ 
    \midrule
    AmazonCounterfactualClassification & 78.69 & 72.45 \\ 
    AmazonPolarityClassification & 95.91 & 94.3 \\ 
    AmazonReviewsClassification & 55.79 & 55.46 \\ 
    Banking77Classification & 88.23 & 87.33 \\ 
    EmotionClassification & 49.77 & 74 \\ 
    ImdbClassification & 94.78 & 92.52 \\ 
    MassiveIntentClassification & 80.57 & 75.64 \\ 
    MassiveScenarioClassification & 82.39 & 78 \\ 
    MTOPDomainClassification & 96.12 & 96.81 \\ 
    MTOPIntentClassification & 86.11 & 87.34 \\ 
    ToxicConversationsClassification & 69.59 & 82.84 \\ 
    TweetSentimentExtractionClassification & 63.72 & 72.74 \\ 
    SprintDuplicateQuestions & 95.66 & 90.99 \\ 
    TwitterSemEval2015 & 81.62 & 68.49 \\ 
    TwitterURLCorpus & 87.75 & 85.35 \\ 
    ArxivClusteringP2P & 50.45 & 35.6 \\ 
    ArxivClusteringS2S & 45.5 & 22.25 \\ 
    BiorxivClusteringP2P & 43.53 & 39.93 \\ 
    BiorxivClusteringS2S & 40.24 & 29.3 \\ 
    MedrxivClusteringP2P & 38.19 & 41.2 \\ 
    MedrxivClusteringS2S & 37.45 & 35.53 \\ 
    RedditClustering & 57.71 & 39.94 \\ 
    RedditClusteringP2P & 66.49 & 53.4 \\ 
    StackExchangeClustering & 73.1 & 46.41 \\ 
    StackExchangeClusteringP2P & 45.91 & 39.7 \\ 
    TwentyNewsgroupsClustering & 54.31 & 38.5 \\ 
    AskUbuntuDupQuestions & 66.98 & 60.56 \\ 
    MindSmallReranking & 32.6 & 24.55 \\ 
    SciDocsRR & 86.33 & 34.94 \\ 
    StackOverflowDupQuestions & 54.91 & 46.04 \\ 
    ArguAna & 61.88 & 74.15 \\ 
    ClimateFEVER & 38.4 & 29.24 \\ 
    CQADupstackTexRetrieval & 42.97 & 23.22 \\ 
    DBPedia & 48.9 & 17.98 \\ 
    FEVER & 87.8 & 82.77 \\ 
    FiQA2018 & 56.62 & 14.91 \\ 
    HotpotQA & 75.7 & 49.04 \\ 
    MSMARCO & 43.1 & 56.43 \\ 
    NFCorpus & 38.59 & 5.48 \\ 
    NQ & 63.5 & 42.95 \\ 
    QuoraRetrieval & 89.62 & 89.1 \\ 
    SCIDOCS & 16.27 & 5.53 \\ 
    SciFact & 76.41 & 66.09 \\ 
    Touche2020 & 26.39 & 6.33 \\ 
    TRECCOVID & 87.33 & 18.22 \\ 
    STS12 & 79.65 & 74.26 \\ 
    STS13 & 88.43 & 84.2 \\ 
    STS14 & 84.54 & 77.5 \\ 
    STS15 & 90.42 & 84.95 \\ 
    STS16 & 87.68 & 82.21 \\ 
    STS17 & 91.75 & 81.67 \\ 
    STS22 & 67.28 & 71.25 \\ 
    BIOSSES & 82.64 & 84.22 \\ 
    SICK-R & 80.76 & 78 \\ 
    STSBenchmark & 88.6 & 84.18 \\ 
    SummEval & 31.4 & 32.36 \\ 
    \midrule
    Avg. & 67.69 & 57.20 \\
    \bottomrule
  \end{tabular}
  }
  \caption{Detailed results of E5-Mistral and LUSIFER on the English benchmark datasets.}
\end{table}

\begin{table}[!htp]
    \centering
    \small
    \resizebox{\columnwidth}{!}{
    \begin{tabular}{l|c|c}
      \toprule
        Ru Datasets & E5-Mistral & LUSIFER \\ 
      \midrule
        GeoreviewClassification & 46.92 & 43.79 \\ 
        HeadlineClassification & 76.52 & 79.26 \\ 
        InappropriatenessClassification & 59.35 & 63.15 \\ 
        KinopoiskClassification & 60.67 & 60.57 \\ 
        MassiveIntentClassification & 72.06 & 71.29 \\ 
        MassiveScenarioClassification & 76.64 & 74.49 \\ 
        RuReviewsClassification & 64.10 & 67.40 \\ 
        RuSciBenchGRNTIClassification & 60.19 & 59.51 \\ 
        RuSciBenchOECDClassification & 46.30 & 46.41 \\ 
        GeoreviewClusteringP2P & 69.87 & 59.20 \\ 
        RuSciBenchGRNTIClusteringP2P & 52.96 & 55.00 \\ 
        RuSciBenchOECDClusteringP2P & 46.54 & 49.95 \\ 
        TERRa & 57.45 & 54.24 \\ 
        RiaNewsRetrieval & 71.39 & 49.61 \\ 
        RuBQRetrieval & 38.04 & 43.48 \\ 
        RuSTSBenchmarkSTS & 81.79 & 78.20 \\ 
        STS22 & 61.32 & 61.44 \\ 
      \midrule
        Avg. & 61.30 & 59.82 \\ 
      \bottomrule
    \end{tabular}
    }
    \caption{Detailed results of E5-Mistral and LUSIFER on the Russian benchmark datasets.}
\end{table}

\begin{table}[!ht]
  \centering
  \small
  \resizebox{\columnwidth}{!}{
  \begin{tabular}{l|c|c}
    \toprule
    Fr Datasets & E5-Mistral & LUSIFER \\ 
    \midrule
    AmazonReviewsClassification & 43.36 & 49.96 \\ 
    MTOPIntentClassification & 70.39 & 79.14 \\ 
    MassiveIntentClassification & 71.12 & 70.88 \\ 
    MassiveScenarioClassification & 74.68 & 73.96 \\ 
    TweetSentimentClassification & 50.23 & 62.62 \\ 
    SIB200Classification & 72.45 & 79.51 \\ 
    FrenchBookReviews & 46.77 & 48.07 \\ 
    PawsXPairClassification & 62.15 & 65.93 \\ 
    RTE3 & 88.45 & 87.62 \\ 
    XNLI & 76.60 & 62.75 \\ 
    MasakhaNEWSClusteringP2P & 50.96 & 48.59 \\ 
    MasakhaNEWSClusteringS2S & 52.08 & 63.12 \\ 
    MLSUMClusteringP2P & 42.69 & 42.70 \\ 
    MLSUMClusteringS2S & 42.60 & 41.51 \\ 
    HALClusteringS2S & 24.21 & 24.16 \\ 
    SIB200ClusteringS2S & 29.94 & 43.30 \\ 
    MultiEURLEXMultilabelClassification & 5.00 & 3.51 \\ 
    BelebeleRetrieval & 84.66 & 83.76 \\ 
    MintakaRetrieval & 52.60 & 18.88 \\ 
    OpusparcusPC & 94.58 & 90.63 \\ 
    STS17 & 84.66 & 82.19 \\ 
    SICKFr & 79.12 & 74.22 \\ 
    STS22 & 76.50 & 73.77 \\ 
    STSBenchmarkMultilingualSTS & 83.98 & 78.42 \\ 
    SummEvalFr & 31.38 & 31.91 \\ 
    \midrule
    Avg. & 59.65 & 59.24 \\ 
    \bottomrule
  \end{tabular}
  }
  \caption{Detailed results of E5-Mistral and LUSIFER on the French benchmark datasets.}
\end{table}

\begin{table}[!ht]
  \centering
  \small
  \resizebox{\columnwidth}{!}{
  \begin{tabular}{l|c|c}
    \toprule
    Vi Datasets & E5-Mistral & LUSIFER \\ 
    \midrule
    MassiveIntentClassification & 66.36 & 71.38 \\ 
    MassiveScenarioClassification & 70.69 & 74.82 \\ 
    MultilingualSentimentClassification & 69.30 & 81.30 \\ 
    SIB200Classification & 70.20 & 78.58 \\ 
    VieStudentFeedbackClassification & 73.02 & 77.39 \\ 
    XNLI & 71.32 & 61.30 \\ 
    SIB200ClusteringS2S & 32.93 & 46.79 \\ 
    BelebeleRetrieval & 79.20 & 85.51 \\ 
    MLQARetrieval & 32.43 & 54.61 \\ 
    VieQuADRetrieval & 20.35 & 45.20 \\ 
    \midrule
    Avg. & 58.58 & 67.69 \\ 
    \bottomrule
  \end{tabular}
  }
  \caption{Detailed results of E5-Mistral and LUSIFER on the Vietnamese benchmark datasets.}
\end{table}

\begin{table}[!ht]
  \centering
  \small
  \resizebox{\columnwidth}{!}{
  \begin{tabular}{l|c|c}
    \toprule
    Fa Datasets & E5-Mistral & LUSIFER \\ 
    \midrule
    MassiveScenarioClassification & 76.37 & 77.94 \\ 
    MassiveIntentClassification & 71.98 & 73.32 \\ 
    MultilingualSentimentClassification & 80.07 & 80.54 \\ 
    FarsTail & 63.49 & 67.98 \\ 
    WikipediaRerankingMultilingual & 75.60 & 78.75 \\ 
    WikipediaRetrievalMultilingual & 67.77 & 78.49 \\ 
    \midrule
    Avg. & 72.55 & 76.17 \\ 
    \bottomrule
  \end{tabular}
  }
  \caption{Detailed results of E5-Mistral and LUSIFER on the Farsi benchmark datasets.}
\end{table}

\begin{table}[!ht]
  \centering
  \small
  \resizebox{\columnwidth}{!}{
  \begin{tabular}{l|c|c}
    \toprule
    Id Datasets & E5-Mistral & LUSIFER \\ 
    \midrule
    IndonesianMongabayConservationClassification & 24.72 & 25.27 \\ 
    MassiveIntentClassification & 69.51 & 71.38 \\ 
    MassiveScenarioClassification & 72.89 & 74.62 \\ 
    SIB200Classification & 80.88 & 80.44 \\ 
    indonli & 50.00 & 50.22 \\ 
    SIB200ClusteringS2S & 46.46 & 47.50 \\ 
    BelebeleRetrieval & 81.10 & 87.56 \\ 
    SemRel24STS & 40.40 & 40.57 \\ 
    \midrule
    Avg. & 58.25 & 59.70 \\ 
    \bottomrule
  \end{tabular}
  }
  \caption{Detailed results of E5-Mistral and LUSIFER on the Indonesian benchmark datasets.}
\end{table}

\begin{table}[!ht]
  \centering
  \small
  \resizebox{\columnwidth}{!}{
  \begin{tabular}{l|c|c}
    \toprule
    Ar Datasets & E5-Mistral & LUSIFER \\ 
    \midrule
    TweetEmotionClassification & 53.74 & 49.03 \\ 
    ArEntail & 77.63 & 84.15 \\ 
    XNLI & 68.00 & 58.58 \\ 
    MintakaRetrieval & 17.15 & 16.59 \\ 
    MLQARetrieval & 28.32 & 47.90 \\ 
    STS17 & 75.13 & 71.44 \\ 
    STS22 & 61.01 & 61.54 \\ 
    \midrule
    Avg. & 54.43 & 55.60 \\ 
    \bottomrule
  \end{tabular}
  }
  \caption{Detailed results of E5-Mistral and LUSIFER on the Arabic benchmark datasets.}
\end{table}

\begin{table}[!ht]
  \centering
  \small
  \resizebox{\columnwidth}{!}{
  \begin{tabular}{l|c|c}
    \toprule
    Fi Datasets & E5-Mistral & LUSIFER \\ 
    \midrule
    FinToxicityClassification & 53.78 & 62.23 \\ 
    MassiveIntentClassification & 64.15 & 70.77 \\ 
    MassiveScenarioClassification & 67.79 & 75.02 \\ 
    MultilingualSentimentClassification & 72.42 & 83.59 \\ 
    SIB200Classification & 66.57 & 77.06 \\ 
    WikipediaRerankingMultilingual & 86.85 & 82.65 \\ 
    BelebeleRetrieval & 73.89 & 85.18 \\ 
    WikipediaRetrievalMultilingual & 71.90 & 82.94 \\ 
    OpusparcusPC & 91.41 & 91.63 \\ 
    FinParaSTS & 20.97 & 17.24 \\ 
    \midrule
    Avg. & 66.97 & 72.83 \\ 
    \bottomrule
  \end{tabular}
  }
  \caption{Detailed results of E5-Mistral and LUSIFER on the Finnish benchmark datasets.}
\end{table}

\begin{table}[!ht]
  \centering
  \small
  \resizebox{\columnwidth}{!}{
  \begin{tabular}{l|c|c}
    \toprule
    Ko Datasets & E5-Mistral & LUSIFER \\ 
    \midrule
    MassiveIntentClassification & 70.42 & 69.79 \\ 
    MassiveScenarioClassification & 75.12 & 75.60 \\ 
    KorSarcasmClassification & 57.64 & 55.28 \\ 
    SIB200Classification & 72.70 & 77.89 \\ 
    KorHateSpeechMLClassification & 8.49 & 7.54 \\ 
    PawsXPairClassification & 53.10 & 54.97 \\ 
    KLUE-TC & 60.58 & 63.95 \\ 
    SIB200ClusteringS2S & 31.04 & 46.58 \\ 
    Ko-StrategyQA & 63.81 & 68.66 \\ 
    BelebeleRetrieval & 80.09 & 84.69 \\ 
    KLUE-STS & 83.48 & 84.17 \\ 
    KorSTS & 79.28 & 78.36 \\ 
    STS17 & 80.97 & 80.55 \\ 
    \midrule
    Avg. & 62.82 & 65.23 \\ 
    \bottomrule
  \end{tabular}
  }
  \caption{Detailed results of E5-Mistral and LUSIFER on the Korean benchmark datasets.}
\end{table}

\begin{table}[!ht]
  \centering
  \small
  \resizebox{\columnwidth}{!}{
  \begin{tabular}{l|c|c}
    \toprule
    Hi Datasets & E5-Mistral & LUSIFER \\ 
    \midrule
    MTOPIntentClassification & 68.84 & 79.93 \\ 
    SentimentAnalysisHindi & 58.98 & 73.92 \\ 
    MassiveIntentClassification & 64.69 & 71.01 \\ 
    MassiveScenarioClassification & 69.71 & 75.42 \\ 
    SIB200Classification & 68.43 & 75.98 \\ 
    TweetSentimentClassification & 37.70 & 40.78 \\ 
    XNLI & 65.04 & 60.26 \\ 
    IndicReviewsClusteringP2P & 40.04 & 42.40 \\ 
    SIB200ClusteringS2S & 27.32 & 45.62 \\ 
    WikipediaRerankingMultilingual & 85.22 & 78.17 \\ 
    BelebeleRetrieval & 69.73 & 66.76 \\ 
    MintakaRetrieval & 18.60 & 21.53 \\ 
    MLQARetrieval & 35.37 & 54.54 \\ 
    WikipediaRetrievalMultilingual & 74.62 & 75.25 \\ 
    IndicCrosslingualSTS & 42.30 & 58.97 \\ 
    SemRel24STS & 73.14 & 77.34 \\ 
    \midrule
    Avg. & 56.23 & 62.37 \\ 
    \bottomrule
  \end{tabular}
  }
  \caption{Detailed results of E5-Mistral and LUSIFER on the Hindi benchmark datasets.}
\end{table}

\begin{table}[!ht]
  \centering
  \small
  \resizebox{\columnwidth}{!}{
  \begin{tabular}{l|c|c}
    \toprule
    Bn Datasets & E5-Mistral & LUSIFER \\ 
    \midrule
    BengaliDocumentClassification & 50.78 & 48.00 \\ 
    BengaliHateSpeechClassification & 54.67 & 51.43 \\ 
    MassiveIntentClassification & 59.51 & 66.65 \\ 
    MassiveScenarioClassification & 64.57 & 70.91 \\ 
    XNLIV2 & 63.66 & 60.01 \\ 
    IndicReviewsClusteringP2P & 38.20 & 45.68 \\ 
    SIB200ClusteringS2S & 23.88 & 43.96 \\ 
    WikipediaRerankingMultilingual & 82.66 & 76.39 \\ 
    BelebeleRetrieval & 60.17 & 55.77 \\ 
    IndicQARetrieval & 56.59 & 68.06 \\ 
    WikipediaRetrievalMultilingual & 71.05 & 72.47 \\ 
    IndicCrosslingualSTS & 35.42 & 41.86 \\ 
    \midrule
    Avg. & 55.10 & 58.43 \\ 
    \bottomrule
  \end{tabular}
  }
  \caption{Detailed results of E5-Mistral and LUSIFER on the Bengali benchmark datasets.}
\end{table}

\begin{table}[t]
  \centering
  \small
  \resizebox{\columnwidth}{!}{
  \begin{tabular}{l|c|c}
    \toprule
    Te Datasets & E5-Mistral & LUSIFER \\ 
    \midrule
    IndicNLPNewsClassification & 89.46 & 98.90 \\ 
    IndicSentimentClassification & 61.53 & 90.63 \\ 
    MassiveIntentClassification & 47.34 & 68.69 \\ 
    MassiveScenarioClassification & 51.67 & 74.17 \\ 
    SIB200Classification & 46.23 & 74.56 \\ 
    TeluguAndhraJyotiNewsClassification & 67.40 & 76.24 \\ 
    IndicReviewsClusteringP2P & 34.02 & 43.62 \\ 
    SIB200ClusteringS2S & 10.81 & 42.11 \\ 
    BelebeleRetrieval & 42.46 & 80.32 \\ 
    IndicQARetrieval & 33.67 & 57.61 \\ 
    IndicCrosslingualSTS & 8.36 & 43.76 \\ 
    SemRel24STS & 72.83 & 80.99 \\ 
    \midrule
    Avg. & 47.15 & 69.30 \\ 
    \bottomrule
  \end{tabular}
  }
  \caption{Detailed results of E5-Mistral and LUSIFER on the Telugu benchmark datasets.}
\end{table}

\begin{table}[t]
  \centering
  \small
  \resizebox{\columnwidth}{!}{
  \begin{tabular}{l|c|c}
    \toprule
    Sw Datasets & E5-Mistral & LUSIFER \\ 
    \midrule
    AfriSentiClassification & 39.67 & 46.47 \\ 
    MasakhaNEWSClassification & 72.96 & 74.79 \\ 
    MassiveIntentClassification & 52.84 & 52.79 \\ 
    MassiveScenarioClassification & 61.09 & 58.59 \\ 
    SwahiliNewsClassification & 63.95 & 61.56 \\ 
    XNLI & 58.86 & 57.82 \\ 
    MasakhaNEWSClusteringP2P & 34.15 & 36.95 \\ 
    MasakhaNEWSClusteringS2S & 21.34 & 35.97 \\ 
    \midrule
    Avg. & 50.61 & 53.12 \\ 
    \bottomrule
  \end{tabular}
  }
  \caption{Detailed results of E5-Mistral and LUSIFER on the Swahili benchmark datasets.}
\end{table}

\end{document}